\documentclass[final]{nesy2025-arxiv} 

\usepackage{longtable}
\usepackage{rotating}  
\usepackage{booktabs}
\usepackage{wrapfig}
\usepackage{booktabs}
\usepackage[load-configurations=version-1]{siunitx} 
\usepackage[cachedir=.]{minted}
\usepackage{xcolor}
\usepackage{listings}
\usepackage{adjustbox}
\usepackage{amssymb}
\usepackage{pifont}
\usepackage{multirow}
\usepackage{graphicx}   
\usepackage{subcaption} 

\newcommand{\xmark}{\ding{55}}

\theorembodyfont{\upshape}
\theoremheaderfont{\scshape}
\theorempostheader{:}
\theoremsep{\newline}

\usepackage{array}
\newcolumntype{P}[1]{>{\raggedright\arraybackslash}p{#1}}

\title[NeSy Frameworks]{Neuro-Symbolic Frameworks: Conceptual Characterization and Empirical Comparative Analysis}

\newcommand{\s}[1]{{\textcolor{black}{#1}}}
\newcommand{\sr}[1]{{\textcolor{black}{#1}}}

 \author{\Name{Sania Sinha} \Email{sinhasa3@msu.edu}\\
  \Name{Tanawan Premsri} \Email{premsrit@msu.edu} \\
  \Name{Danial Kamali} \Email{kamalida@msu.edu}  \\
  \Name{Parisa Kordjamshidi} \Email{kordjams@msu.edu}\\
  \addr Department of Computer Science and Engineering, Michigan State University}

\begin{document}

\maketitle

\begin{abstract}
  Neurosymbolic (NeSy) frameworks combine neural representations and learning with symbolic representations and reasoning. Combining the reasoning capacities, explainability, and interpretability of symbolic processing with the flexibility and power of neural computing allows us to solve complex problems with more reliability while being data-efficient. However, this recently growing topic poses a challenge to developers with its learning curve, lack of user-friendly tools, libraries, and unifying frameworks. In this paper, we  characterize the technical facets of existing NeSy frameworks, such as the symbolic representation language, integration with neural models, and the underlying algorithms. A majority of the NeSy research focuses on algorithms instead of providing generic frameworks for declarative problem specification to leverage  problem solving. To highlight the key aspects of Neurosymbolic modeling, we showcase three generic NeSy frameworks - \textit{DeepProbLog}, \textit{Scallop}, and \textit{DomiKnowS}. We identify the challenges within each facet that lay the foundation for identifying the expressivity of each framework in solving a variety of problems. Building on this foundation, we aim to spark transformative action and encourage the community to rethink this problem in novel ways.
\end{abstract}


\section{Introduction}
\label{sec:intro}

\textbf{Symbolic or \textit{good old-fashioned} AI} focused on creating rule-based reasoning systems~\citep{hayes1985rule} exemplified with early works  such as the Physical Symbol System~\citep{Augusto2021-AUGFST-2,NEWELL1980135} and ELIZA~\citep{10.1145/365153.365168}. However, drawbacks such as limited scalability due to the need to explicitly define domain predicates and rules for each task, lack of robustness in handling messy real-world data, and low computational efficiency led to a decline in the popularity of this paradigm, shifting the focus toward neural computing and deep learning. 
\textbf{Deep Learning}~\citep{DL, Ahmad2019} revolutionized AI as nuanced relationships in data could be learned by backpropagation through multiple layers of processing and creating abstract representations of data. However, it led to a loss of explainability~\citep{10.1145/3555803}, dependence on large amounts of data, and rising concerns about its environmental sustainability~\citep{10.1145/3442188.3445922}. \textbf{Neurosymbolic AI}~\citep{hitzler2022neuro, Bhuyan2024}, a combination of symbolic AI and reasoning with neural networks, attempts to incorporate the capabilities of both worlds and create systems that are data and time efficient, generalizable, and explainable. 
Neurosymbolic models have been applied to several real-world applications~\citep{bouneffouf2022surveyapplicationsneurosymbolicartificial} in safety-critical areas~\citep{rel} such as healthcare~\citep{hossain2025studyneurosymbolicartificialintelligence} and autonomous driving~\citep{pmlr-v155-sun21a}.
Several techniques have been proposed for this integration \citep{Kautz_2022, 10981497}, trying to combine the pros and mitigate the cons from both symbolic and neural methods.
However, due to lack of unified libraries to facilitate this research and the focus on specific algorithms rather than generic frameworks, this research becomes less impactful. Moreover, the few generic frameworks tend to vary in problem formulation, implementation, algorithms, and flexibility of use. This poses a challenge in being able to compare their performance uniformly or identify a research direction that improves on previous work. To alleviate this issue, we provide a comparative study with the following key contributions. 

a) Identifying the main components of existing NeSy frameworks,
b) Comparison of frameworks across the identified facets,
c) Highlighting the requirements for the next generation of NeSy frameworks, building upon the drawbacks of the current systems and the possible interplays between the neural and symbolic components. 
We plan to expand this study to cover more frameworks, while the three selected ones are used to explain the aspects of our characterization. These frameworks are demonstrated with four example tasks detailed in Section~\ref{sec:tasks}, tying the comparative facets concretely with a technical implementation.\footnote{\url{https://github.com/HLR/nesy-examples}} 

\begin{figure*}[hbt!]
    \centering
    \includegraphics[width=0.9\textwidth]{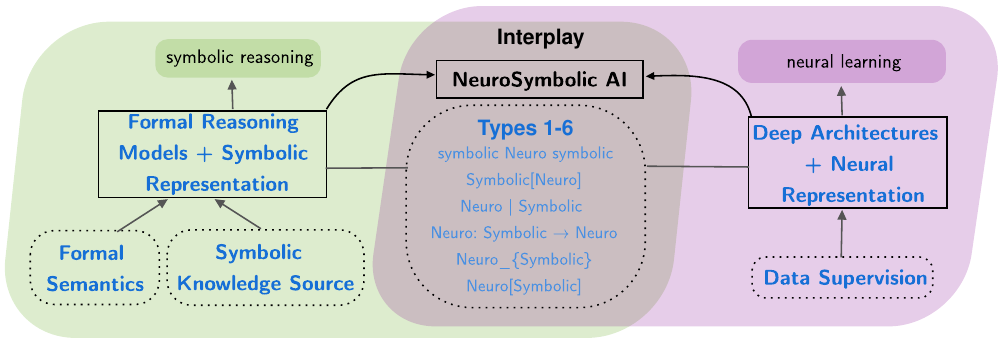}
    \caption{An overview of the main components  of a neurosymbolic framework.}
    \label{fig:overview}
\end{figure*}

\vspace{-2mm}
\section{Neurosymbolic Frameworks}
\label{sec:frameworks}
A NeSy framework should provide flexibility for modeling both neural and symbolic components~\citep{kordjamshidi-etal-2016-better, kjam} and their interplay in a unified declarative framework, going beyond specific underlying algorithms and techniques~\cite{kjam}. 
On the symbolic side, a generic framework should support a symbolic representation language that can be seamlessly connected to neural components and cover different symbolic reasoning mechanisms. On the neural side, we need to have the flexibility of connecting to various architectures, including various loss functions, sources of supervision, and training paradigms. More importantly, a NeSy framework should provide a modeling language for specification and seamless integration of the two components in building pipelines or arbitrary composition of models. Such a NeSy framework should support neuro-symbolic training and inference beyond specific integration algorithms. 
We distinguish between \textit{NeSy techniques} and \textit{NeSy frameworks}. By techniques, we mean when task-specific solutions are provided~\citep{Lample2020Deep, 1181487}. For example, \textbf{AlphaGo}~\citep{alphago} introduced a reinforcement learning solution to Go, using Monte Carlo Tree Search as a symbolic component inside a neural network. Another example is \textbf{NS-CL}~\citep{mao2019neurosymbolicconceptlearnerinterpreting} (Neuro-Symbolic Concept Learner) that integrates neural perception with symbolic reasoning to learn visual concepts and compositional language grounding for VQA tasks.
Many other techniques and algorithms are proposed for the interplay between the two paradigms~\citep{BADREDDINE2022103649, cohen2017tensorlog, tensor, inproceedings, sathasivam2011learning, serafini2016learning, lamb2021graphneuralnetworksmeet} such as Inference Masked Loss~\citep{ijcai2020p382}, Semantic Loss~\citep{pmlr-v80-xu18h}, Primal-Dual~\citep{NEURIPS2019_cf708fc1}, etc., later discussed in Section \ref{sec: interplay}.
NeSy techniques often lack the generality of frameworks, which are designed as broader tools intended for practical use and extensibility with new integration algorithms {and with the capability of programming and configuring the two parts and their interplay.} 

In this work, we focus on a selection of generic NeSy frameworks. The following are examples of research efforts towards advancing the development of such general-purpose frameworks:
\textbf{DeepProbLog}~\citep{MANHAEVE2021103504} is a probabilistic logic programming language, incorporating neural predicates in logic programming with an underlying differentiable translation of logical reasoning. The probabilistic logic programming component is built on top of ProbLog~\citep{10.5555/1625275.1625673}.
\textbf{DomiKnowS}~\citep{rajaby-faghihi-etal-2021-domiknows, faghihi2023glueconsgenericbenchmarklearning, faghihi2024prompt2demodeldeclarativeneurosymbolicmodeling} is a declarative learning-based programming framework~\citep{Kordjamshidi2019DeclarativeLP} that integrates symbolic domain knowledge into deep learning. It is a Python framework, facilitating the incorporation of logical constraints that represent domain knowledge with neural learning in PyTorch.
\textbf{Scallop}~\citep{NEURIPS2021_d367eef1, 10.1145/3591280, Li_2024} is a framework that includes flexible symbolic representation based on relational data modeling, using a declarative logic programming built on top of Datalog~\citep{10.5555/551350} with a framework for automatic differentiable reasoning.
\textbf{LEFT}~\citep{hsu2023whatsleftconceptgrounding} is a less generic framework designed for grounding language in visual modality and compositional reasoning over concepts. The framework consists of an LLM interpreter that converts natural language to logical programs. The generated programs are directed to a differentiable, domain-independent, and soft first-order logic-based executor. LEFT is limited to tasks requiring grounding language in vision such as visual question answering~\citep{Johnson_2017_CVPR, yi2018neural, Liu_2019_CVPR}. Building on this foundation, NeSyCoCo~\citep{nesycoco} was introduced to address the limitations of LEFT, particularly its struggle with lexical variety and handling unseen concepts. NeSyCoCo extends LEFT's approach by using distributed word representations to connect a wide variety of linguistically motivated predicates to neural modules, thus alleviating the reliance on a predefined predicate vocabulary. \textbf{PyReason}~\citep{aditya2023pyreasonsoftwareopenworld} is a library built to support reasoning on top of outputs from neural networks. The neural component produces outputs such as labels or concept scores. While the symbolic component does graph-based reasoning using logic rules declared over a graph structure.
It can produce an explanation trace for inference and has a memory-efficient implementation.
\textbf{PLoT}~\citep{wong2023wordmodelsworldmodels} 
(Probabilistic Language of Thought) is a \textit{proposed} framework leveraging neural and probabilistic modeling for generative world modeling. It models thinking with probabilistic programs and meaning construction with neural programs. The goal is to provide a language-driven unified thinking interface.
\s{\textbf{CCN+}~\citep{GIUNCHIGLIA2024109124} is a framework that modifies the output layer of a neural network to make results compliant with requirements that can be expressed in propositional logic. A requirement layer, ReqL, is built on top of the neural network. The standard cross-entropy loss is adapted into a ReqLoss to learn from the constraints in the ReqL layer.}
\textbf{DeepLog} \citep{derkinderen2025deeplogneurosymbolicmachine} is another \textit{proposed} neurosymbolic AI framework that unifies logic and neural computation under a declarative paradigm. It introduces a DeepLog language, which is an annotated neural extension of grounded first-order logic capable of abstracting various logics and applying them either in model architecture or loss functions. It employs computational algebraic circuits implemented on GPUs, forming a neurosymbolic abstract machine. Together, DeepLog allows efficient specification and execution of diverse neurosymbolic models and inference tasks in a declarative fashion.

We characterize frameworks based on: a) Symbolic knowledge representation language, b) Representation and flexibility of Neural Modeling, c) Model Declaration, d) Interplay between symbolic and sub-symbolic systems, and e) The usage of LLMs. 
Figure \ref{fig:overview} shows the relationship between these different aspects. The neural representations and the symbolic representations are the two main components of a neurosymbolic framework. The neural representation guides learning and obtaining supervision from the data, while the symbolic representations leverage symbolic reasoning, where the  symbolic knowledge can be exploited during training or inference. 
\s{Table \ref{tab:frameworks} shows an overview of the frameworks across chosen features. For future sections, we focus on \textbf{DomiKnowS}, \textbf{DeepProbLog}, and \textbf{Scallop} to provide a deeper investigation of the challenges in each component. Due to differences in implementation, each framework allows for easy implementation of different types of tasks. The chosen frameworks enable us to solve the same task in multiple frameworks.}

\begin{table*}
    \centering
    \small
    \setlength{\tabcolsep}{1mm}
    \begin{tabular}{P{0.14\textwidth}P{0.09\textwidth}P{0.19\textwidth} >{\centering\arraybackslash}P{0.10\textwidth}P{0.15\textwidth}>{\centering\arraybackslash}P{0.05\textwidth}P{0.13\textwidth}}
        \toprule
        \multirow{2}{*}{\textbf{Framework}} & \multicolumn{2}{c}{\textbf{Symbolic}}  & \multirow{2}{=}{\textbf{Model Dec}} & \multicolumn{2}{c}{\textbf{Interplay}} & \multirow{2}{*}{\textbf{LLM}} \\
        \cline{2-3} \cline{5-6}
        & \textbf{Lang} & \textbf{Knowledge Rep} &  &\textbf{Algorithm} & \textbf{Eff} & \\
        \hline
        CCN+ & None & Propositional Logic Clauses & \xmark & ReqL \& ReqLoss & \xmark & \xmark \\
        \hline
        DomiKnowS & None & Concepts, Constraints & \checkmark & Primal-Dual, Sampling Loss & \xmark & \cite{faghihi2024prompt2demodeldeclarativeneurosymbolicmodeling} \\
        \hline
        DeepProbLog& ProbLog & Facts, Rules, Predicates & \xmark & Entailment & \xmark & \xmark \\
        \hline
        LEFT & None & First Order Logic & \xmark & Differentiable Reasoning & \xmark & \cite{hsu2023whatsleftconceptgrounding} \\
        \hline
        PyReason & None & Constants, Relations, Facts, Rules  & \xmark & Reasoning over graph & \checkmark & \xmark \\
        \hline
        Scallop & DataLog & Rules, Relations & \xmark & Differentiable Reasoning & \checkmark & \cite{Li_2024} \\
        \bottomrule
    \end{tabular}
    \caption{Frameworks with their comparative factors. Lang: External language required, Knowledge Rep: Knowledge Representation, Model Dec: Model Declaration flexibility, Algorithm: Supported algorithm(s) for learning and inference, Eff: Computational efficiency considerations, LLM: Use of Large Language Models. }
    \label{tab:frameworks}
\end{table*}

\vspace{-2mm}
\section{Symbolic Knowledge Representation} 
\label{sec:symbolic}

\begin{wrapfigure}{r}{0.5\linewidth}
    \centering
    \includegraphics[width=1\linewidth]{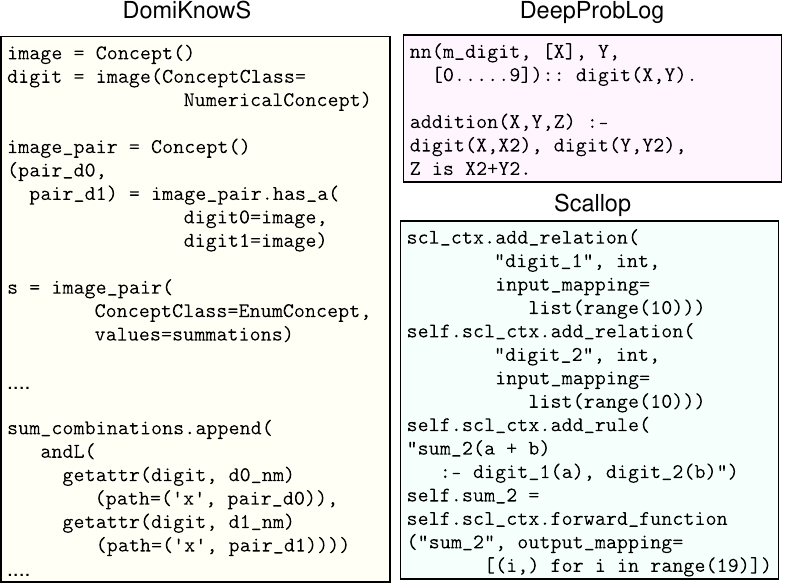}
    \caption{Comparison of Symbolic Representation across frameworks.  }
    \label{fig:sym}
\end{wrapfigure}

Generic Neuro-Symbolic (NeSy) systems and frameworks use symbolic knowledge representation languages to encode constraints, facts, probabilities, and rules. Frameworks vary in how they represent and integrate this symbolic knowledge. Many employ classical formal logic, grounded in well-defined syntax and semantics, and adapt these representations and reasoning mechanisms within a unified integration framework. Some frameworks build on established formalisms such as logic programming or constraint satisfaction. In contrast, others take an entirely new hybrid semantics, while preserving conventional symbolic syntax. 
Figure \ref{fig:sym} compares the implementation of symbolic knowledge (concepts or facts) for the MNIST Sum task. In general, the domain knowledge consists of the two concepts of \textit{digits} and the \textit{sum}.
As can be seen, DomiKnowS represents a part of symbolic domain knowledge as a graph $G(V,E)$, where the nodes are the concepts in the domain and the edges are the relationships between them. Each node can have properties. More complex knowledge beyond entities and relations is expressed with a pseudo first-order logical language with quantifiers designed in Python. DomiKnowS mostly interprets the symbolic knowledge as logical constraints, such as the implementation of \texttt{sum\_combinations} in the given example. Unlike the other frameworks, DomiKnowS does not build on predefined formal semantics. It follows a FOL-like syntax for symbolic logical representations, making it independent of the formal semantics of an underlying formal language and allows more flexibility of representations and adaptation to underlying algorithms in the framework.
DeepProbLog, on the other hand, utilizes logical predicates that are originally a part of the probabilistic logic programs~\citep{NG1992150} of ProbLog~\citep{10.5555/1625275.1625673}, for its symbolic representation. These neural predicates obtain their probability distributions from the underlying neural models. Probabilistic facts, neural facts, and neural annotated disjunctions (nAD) whose probabilities are supplied by the neural component of the program can be added. 
Here, \texttt{digit} is a neural predicate as indicated by the use of \texttt{nn(...)}. DeepProbLog follows the formal semantics of Prolog~\citep{clocksin2003programming}, 
followed by ProbLog, its probabilistic extension. 
Finally, Scallop adopts a relational data model for symbolic knowledge representation~\citep{10.1145/298514.298542}. Scallop is built on top of the syntax and formal semantics of Datalog and its probabilistic extensions, relaxing the exact semantics of ProbLog. It allows for the expression of common reasoning, such as aggregation, negation, and recursion.  Similar to DeepProbLog, some of these predicates in the symbolic part obtain their probability distribution from neural models, such as \texttt{digit\_1} and \texttt{digit\_2}. Additionally, while ProbLog requires exhaustive search for computations, Datalog can use top-k results and exploit database optimizations, making Scallop algorithmically more time-efficient than DeepProbLog.

\vspace{-2mm}
\section{Neural Models Representations}

\begin{wrapfigure}{r}{0.5\linewidth}
    \centering
    \includegraphics[width=1\linewidth]{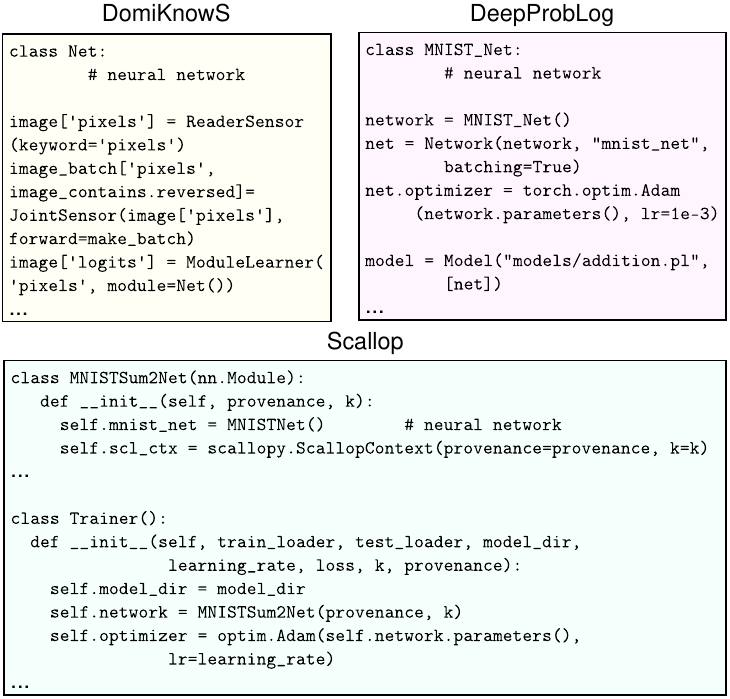}
    \vspace{-3mm}
    \caption{For neural integration, DomiKnowS utilizes sensors and readers for reading in data, while a learner connects to a network. DeepProbLog connects the neural network to the ProbLog file, requiring data handling to construct the terms and queries from the raw data. Scallop has an additional layer on top of the standard network that adds the symbolic context.}
    \label{fig:neural}
\end{wrapfigure}

The other core component of a NeSy system is the neural modeling that is integrated with the symbolic knowledge discussed above. The neural models are mostly wrapped up under the logical predicate names in most of the frameworks that have an explicit logical knowledge representation language. To best leverage the reasoning capabilities of the symbolic system available and the ability of neural models to learn abstract representations from data, the neural models are used as abstract concept learners for the concepts defined as logical predicates in the symbolic representation.
The neural model representation is often used to predict probability distributions for the symbolic concepts based on raw sensory inputs. The neural modeling is often written using standard deep learning libraries, such as  PyTorch~\citep{paszke2019pytorchimperativestylehighperformance}. Figure \ref{fig:neural}  shows snippets of neural modeling
 expressions across frameworks, highlighting differences in implementation. Scallop utilizes relatively standard neural modeling using PyTorch, while needing an added context of symbolic rules. Although integrated into Python, the context relation and rule setup are verbatim from DataLog and only passed as a parameter to a function, which requires familiarity with DataLog and its semantics. DeepProbLog, on the other hand, needs manual configuration of the raw data and processing into queries built for ProbLog, on top of other standard neural components. This processed data is passed into the neural network which is then connected to a ProbLog program, such as \texttt{addition.pl} in the figure. DomiKnowS's neural component is built in PyTorch. Unlike other frameworks, DomiKnowS has built-in components called \textit{Readers,} \textit{Sensors} and \textit{Module learners} that make the connection to neural components \sr{and feeding data to them} explicit in the program. This provides more flexibility in connecting the concepts to deterministic or probabilistic functions that can interact with other symbolic concepts. The module learner can also use custom models. This makes the interaction with raw data structured, transparent, and controllable.

\section{Model Declaration }
\label{sec:composition}

Most frameworks utilize neural components as abstract concept learners and use a symbolic component to reason over the learned concepts. Each learner is a model and \textit{model declaration} refers to the flexibility of modularizing and connecting different learners. 
Each learner can receive supervision independently. 
In most neurosymbolic frameworks, the supervision from data is usually provided based on the final output of the end-to-end model. For example, in an MNIST Sum task used throughout and detailed in Section \ref{sec:tasks}, the neural and symbolic components are trained based on the final output of the sum, without access to individual digit labels in a semi-supervised setting. The task loss, e.g., a Cross-Entropy Loss, is computed, and errors are backpropagated through the differentiable operations that led to the output generation. For example, in DeepProbLog, we can declare a single loss function associated with the entire neural component.
Gradient computations differ across frameworks depending on whether losses are defined individually for each neural output or specified as a single global loss function. 
However, there remains a need for models capable of incorporating supervision at multiple levels of their symbolic representations. 
In DomiKnowS, loss computation can be defined for each symbol. Since each concept is linked to both learning modules and ground-truth labels, their losses can be integrated seamlessly. This enables joint training of all concepts alongside the target task, allowing each concept to be optimized more effectively—leveraging available data without relying solely on the target task's output. In other words, it provides the flexibility of building pipelines of decision making, obtaining distant supervision in addition to joint training and inference.

\vspace{-2mm}
\section{Interplay between Symbolic and Sub-symbolic}
\label{sec: interplay}

\cite{Kautz_2022} provides a characterization of the possible interplays between symbolic and sub-symbolic components. This interplay of neurosymbolic can be explained by the concept of System 1 and System 2 thinking described in \cite{kahneman2011thinking}. Research in this field aims to create an ideal integration that seamlessly supports "thinking fast and slow"~\citep{Booch_Fabiano_Horesh_Kate_Lenchner_Linck_Loreggia_Murgesan_Mattei_Rossi_Srivastava_2021, fabiano2023plansofai}. Here, System 1 refers to the fast neural processing, while System 2 corresponds to the slower, more deliberate symbolic reasoning. \textit{Different methods for the integration of symbolic reasoning and neural programming have been explored such as employing logical constraint satisfaction, integer linear programming, differentiable reasoning, probabilistic logic programming.} 
In this section, we will discuss a system-level algorithmic comparison of the different frameworks.

\textbf{DomiKnowS} models the inference as an integer linear programming problem to enforce the model to follow constraints expressed in first-order logical form~\citep{VANHENTENRYCK1992113}. The objective of the program is guided by the neural components, and the framework supports multiple training algorithms for learning from constraints. The Primal-Dual formulation~\citep{NEURIPS2019_cf708fc1} converts the constrained optimization problem into a min-max optimization with Lagrangian multipliers for each constraint, augmenting the original loss with a soft logic surrogate to minimize constraint violations. Sampling-Loss~\citep{pmlr-v176-ahmed22a}, inspired by semantic loss~\citep{pmlr-v80-xu18h}and samples a set of assignments for each variable based on the probability distribution of the neural modules' output. Integer Linear Programming (ILP)\citep{10.1613/jair.1.13507} formulates an optimization objective based on Inference-Masked Loss\citep{ijcai2020p382} to constrain the model during training. The training goal is to adjust the neural models to produce legitimate outputs that adhere to the given constraints. At prediction time, ILP can also be applied to enforce final predictions that comply with given constraints. DomiKnowS relies on the off-the-shelf optimization solver Gurobi~\citep{gurobi}
\textbf{DeepProbLog} models each problem as a {probabilistic logical} program that consists of neural facts, probabilistic facts, neural predicates, and a set of logical rules. A joint optimization of the parameters of the logic program is done alongside the parameters of the neural component. Neural network training is done using learning from entailment~\citep{frazier1993learning} while in ProbLog, gradient-based optimization is performed on the underlying  generated Arithmetic Circuits~\citep{shpilka2010arithmetic}, which is a differentiable structure. The Arithmetic Circuits are transformed from a Sentential Decision Diagram~\citep{darwiche2011sdd} generated by ProbLog. Algebraic ProbLog~\citep{Kimmig_VandenBroeck_DeRaedt_2011} is used to compute the gradient alongside probabilities using semirings~\citep{eisner2002parameter}.
\textbf{Scallop} is similar in its setup to DeepProbLog where it creates an end-to-end differentiable framework combining a symbolic reasoning component with a neural modeling component. They aim to relax the formal semantics required by the use of ProbLog in DeepProbLog and instead rely on a symbolic reasoning language extending DataLog, built into their framework. They have a customizable provenance semiring framework~\citep{10.1145/1265530.1265535}, where different provenance semirings, such as extended max-min semiring and top-k proofs semiring, allow learning using different types of heuristics for gradient calculations. 
Table~\ref{tab:efficiency} compares the computational efficiency of these models at training and inference time on a single training/testing example. As theoretically suggested, Scallop is expected to outperform other frameworks in inference and training speed, owing to its memory and time-efficient implementation in Rust.
The results in Table~\ref{tab:efficiency} support this expectation, with Scallop achieving the fastest inference time, on par with DomiKnowS.
In practice, DeepProbLog achieves slightly faster training performance than Scallop. 
This discrepancy may be due to overhead unrelated to the core algorithmic complexity.
DomiKnowS exhibits slower training due to the overhead of uploading the entire graph of data into memory.

\vspace{-2mm}
\section{Role of Large Language Models}
\label{sec:llm}

Large foundation models hold significant promise for overcoming the bottleneck of acquiring symbolic representations, which are essential for symbolic reasoning and consequently in neurosymbolic frameworks.

\textbf{Source of Symbolic Knowledge:} The symbolic knowledge in neuro-symbolic systems, which is integrated with the neural component, can originate from several distinct sources. While most systems require explicit, hand-crafted symbolic knowledge, earlier classical logic-based learning research can be used for automatically learning rules from data by using inductive logic programming~\citep{nienhuys1997inductive, 10.1145/219717.219771} or mining constraints. Nowadays, even LLMs can be utilized to generate symbolic knowledge~\citep{pan-etal-2023-logic, mirzaee-kordjamshidi-2023-disentangling, survey_archarya, xu-etal-2024-symbol}.
Several neurosymbolic frameworks and systems have tried utilizing large foundation models to generate the symbolic knowledge, based on the task or query, to overcome the labor-intensive nature of hand-crafting rules for every single task and the time required in the automatic learning of symbolic knowledge from data~\citep{Ishay2023LLM2ASP, xu-etal-2024-symbol, yang-etal-2024-arithmetic}. Extraction of symbolic representations from Foundation Models has become possible given the vast implicit knowledge stored within these models, such as LLMs and multimodal models, which are trained on massive and diverse corpora \citep{Li_2024,petroni2019languagemodelsknowledgebases}. These models can generate symbolic content (e.g., candidate rules, knowledge graph triples, or logic statements), perform reasoning that mimics symbolic inference, or act as components alongside symbolic modules~\citep{fang2024large}.  For example, LLMs can be prompted to extract facts from unstructured text, effectively populating a symbolic knowledge graph \citep{yao2025exploringlargelanguagemodels}. Techniques like Symbolic Chain-of-Thought inject formal logic into the LLM's reasoning process, improving accuracy and explainability on logical reasoning tasks~\citep{xu2024faithful}.
However, foundation models are prone to hallucinations and lack the strict logical guarantees of traditional symbolic systems \citep{zheng2024reliablellmsknowledgebases}. Therefore, integrating foundation models often requires careful prompting, verification steps to ensure reliability~\citep{xu2024faithful}.
\textbf{Generation of inputs to symbolic engines:} LLMs have also been used to generate translations {from raw inputs, specially natural language, to symbolic language that is then} fed into a symbolic reasoner. In examples such as Logic-LM~\citep{pan2023logiclmempoweringlargelanguage}, LLMs are leveraged to convert a natural language query into symbolic language that is then solved by a symbolic reasoner. This method improves the performance of unfinetuned LLMs on logical reasoning-based tasks. DomiKnowS~\citep{faghihi2024prompt2demodeldeclarativeneurosymbolicmodeling} takes this a step further by enabling users to describe problems in natural language which LLMs then use to generate relevant concepts and relationships. Through a user-interactive process, these concepts and relationships are refined iteratively. Finally, the LLM translates the user-defined constraints from natural language into first-order logic representations before converting them into DomiKnowS syntax.  
Some systems use LLMs in multiple capacities. In VIERA~\citep{Li_2024}, which is built on top of Scallop, 12 foundation models can be used as plugins. These models are treated as stateless functions with relational inputs and outputs. These foundation models can be either language models like GPT~\citep{openai2024gpt4technicalreport} and LLaMA~\citep{touvron2023llama2openfoundation}, vision models such as OWL-ViT~\citep{minderer2022simpleopenvocabularyobjectdetection} and SAM~\citep{kirillov2023segment}, or multimodal models such as CLIP~\citep{radford2021learningtransferablevisualmodels}. These models can be used to extract facts, assign probabilities, or for classification, and are treated as "foreign predicates" in their interface. An older version, DSR-LM~\citep{zhang2023improvedlogicalreasoninglanguage} of this utilized BERT-based language models for perception and relation extraction, combined with a symbolic reasoner for question answering. LEFT, on the other hand, uses LLMs both for the generation of the concepts that are used for grounding and as an interpreter to generate the first-order logic program corresponding to a natural language query, that is solved by the symbolic executor. 

\vspace{-2mm}
\section{Example Tasks}
\label{sec:tasks}

NeSy frameworks formulate problems in various ways based on their implementation and symbolic interpretation. In \textbf{DomiKnowS}, the symbolic reasoning part is formulated as a logical constraint solving problem. The domain is represented as a graph $G(V,E)$, where the nodes are the concepts in the domain and the edges are the relationships between them. Each node can have properties. The final logical constraints apply to the graph concepts. In \textbf{DeepProbLog}, the symbolic reasoning problem is interpreted a probabilistic logic programs in ProbLog. 
In \textbf{Scallop}, similarly, the problem is viewed as a combination of the neural and the symbolic components where the symbolic part is a probabilistic logical program similar to DeepProblog with further optimized inference. In \textbf{LEFT}, the problem is limited to the application of concept learning and grounding language into visual modality. Here, the neural model is composed of feature extractors, object and relation classifiers (concept learners), and a first-order logic program generator for a given question. 
In this section, we will compare the problem formulations in each of these frameworks for a set of tasks. 
\textbf{Note} that we only include LEFT for the visual question answering task due to the domain-specific nature of the framework. All code associated with these tasks and referenced in this section is maintained publicly on GitHub. \footnote{\url{https://github.com/HLR/nesy-examples}} 

\begin{table*}[t]
    \centering
    \small
    \setlength{\tabcolsep}{1mm}
    \begin{tabular}{l| p{0.125\textwidth} p{0.125\textwidth} p{0.125\textwidth} p{0.125\textwidth} p{0.125\textwidth} p{0.125\textwidth}}
    \toprule
         & \multicolumn{3}{c}{\textbf{MNIST Sum}} & \multicolumn{3}{c}{\textbf{Toy-NER}} \\
        Framework & Training Time (ms) & Testing Time (ms) & Memory (MB) & Training Time (ms) & Testing Time (ms) & Memory (MB) \\
        \midrule
        DomiKnowS & 37.72 & \textbf{2.34} & \textbf{573.80} & 14.86 & 14.29 & 576.82 \\
        DeepProbLog & \textbf{5.84} & 3.24 & 739.72 & 20.74 & 20.47 & 3767.08 \\
        Scallop & 6.50 & \textbf{2.35} & \textbf{581.36} & \textbf{1.50} & \textbf{1.05} & \textbf{297.1} \\
        
        \midrule
        \midrule
         & \multicolumn{3}{c}{\textbf{Math-Inference}} & \multicolumn{3}{c}{\textbf{Simple VQA}} \\
        Framework & Training Time (ms) & Testing Time (ms) & Memory (MB) & Training Time (ms) & Testing Time (ms) & Memory (MB) \\
        \midrule
        DomiKnowS & 77.46 & 69.58 & 1039.30 & 81.02 & 60.52 & 1413.44 \\
        DeepProbLog & 5.54 & 5.07 & 1588.30 & 755.13  & 363.81 & 1346.23 \\
        Scallop & \textbf{0.948} & \textbf{0.223} & \textbf{345.48
} & \textbf{13.17} & \textbf{22.66} & \textbf{768.4} \\

        LEFT & N/A & N/A & N/A   &6.19 & 3.44 & 755.1 \\
        
    \bottomrule
    \end{tabular}
    \caption{Time and Space efficiency for each framework across 4 tasks. Time and memory records are averaged over 5 runs. Times are in milliseconds per sample.  Memory utilization is in megabytes and indicates the overall memory for training.}
    \label{tab:efficiency}
\end{table*}

\subsection{MNIST Sum}

The MNIST Sum task is an extension of the classic MNIST handwritten digit recognition task~\citep{726791} where given two images of digits, the task is to output their sum that is a whole number. The training examples consist of the two images of the digits and the ground-truth label of their sum. The individual labels of the digits are not available for training.

\subsubsection{DomiKnowS}\label{sec:DomiKnowS_MNIST}

\textbf{Problem Specification.} DomiKnowS formulates the problem using graph representations of concepts, relations, and logic.
For performing the MNIST Sum task in DomiKnowS, the first concept defined is \textit{image} concept representing visual information. The \textit{digit} concept, a subclass of image, is introduced to represent the output class, ranging from 0 to 9.
To establish relationships between digit images, the \textit{image pair} concept is defined as an edge connecting two digit concepts. The sum concept is then introduced under image pair to represent the summation of the two digit concepts and the ground-truth output of the program.
For this task, three constraints are defined.
The first two constraints utilize \textit{exactL} to ensure that the predicted digit and sum values belong to only one valid class. Another constraint enforces that the expected sum value matches the sum of the two digit predictions. This is implemented using \textit{ifL} constraints, which verify whether the predicted digits form one of the possible solutions for a valid sum. If multiple solutions exist, the \textit{orL} constraint ensures that at least one of the answers corresponds to the predicted digits.

\noindent\textbf{Neural Modeling.} 
The model declaration comprises standard neural modeling components, including data loading, pre-processing, neural network definition, and loss function specification.
The process begins with the \textit{ReaderSensor}, which reads the input image. Next, a relation concept is defined using another sensor, \textit{JointSensor}, to establish connections between images. The module learner is then employed to generate an initial prediction for the digit concept, which is subsequently passed to another sensor, \textit{FunctionalSensor}, to compute the sum of two images.

\subsubsection{DeepProbLog}
\textbf{Problem Specification.} 
DeepProbLog formulates a problem regarding probabilistic facts, neural facts, and neural annotated disjunctions (nAD). In the MNIST Sum task, the fact $X$ is defined to represent the input image. A neural network function is then introduced to map $X$ to its corresponding digit, denoted as digit($X, Y$). 
To enforce constraints about the summation and the ground-truth sum, a function is defined to compute the sum of two digits.

\noindent
\textbf{Neural Modeling.}
The neural modeling follows a standard neural network setup, such as a CNN-based classifier. It is preceded by data loading and pre-processing, which are performed separately from the ProbLog program. Thus, the neural model used in DeepProbLog can be initialized independently of the DeepProbLog model.
Once the neural model is initialized, the framework passes it along with a probabilistic program as input.

\subsubsection{Scallop}
\noindent\textbf{Problem Specification.} 
Scallop formulates the problem in terms of relations, values, and (Horn) rules derived from Datalog.
As discussed earlier, the concepts and constraints defined in this framework are similar to those in DeepProbLog.
However, these rules can be directly embedded into a Scallop program through its API.
The process begins by establishing the concepts \textit{digit1} and \textit{digit2} to represent the digit values of two given images. Based on the summation of these two values, it must be equal to the $sum\_2$ logical reasoning module, which serves as the ground truth for this task.

\noindent\textbf{Neural Modeling.}
Unlike DeepProbLog, the neural modeling is integrated with Scallop's relation and rule declaration. The neural modeling remains a standard neural network.

\subsection{Shapes}
The \textbf{Shapes} dataset is a synthetic VQA benchmark designed to evaluate elementary spatial reasoning. Each sample consists of a $128 \times 128$ pixel image where the task is to answer the fixed question: ``Is there a red shape above a blue circle?''. 
The primary rationale for using a synthetic dataset is to create a controlled experimental environment. This approach allows us to isolate specific reasoning skills—such as attribute binding and relational understanding—from the perceptual complexities and spurious correlations.

\begin{figure}[htbp]
  \centering
  \begin{minipage}[b]{0.45\textwidth}
    \centering
    \adjustbox{frame=1pt,margin=2pt}{\includegraphics[width=0.4\textwidth]{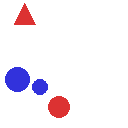}}
    \\[0.5ex]
    \small True example: A red triangle is above a blue circle.
  \end{minipage}%
  \hfill
  \begin{minipage}[b]{0.45\textwidth}
    \centering
    \adjustbox{frame=1pt,margin=2pt}{\includegraphics[width=0.4\textwidth]{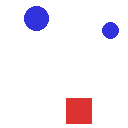}}
    \\[0.5ex]
    \small False example: No red shape is above a blue circle.
  \end{minipage}
  \caption{Examples from the Shapes dataset for the question  
  “Is there a red shape above a blue circle?”}
  \label{fig:shapes_examples}
\end{figure}

Positive examples are generated by programmatically placing a red object (circle, square, or triangle) above a blue circle. In negative examples, this specific spatial configuration is absent. To increase task complexity, every image also contains one to three randomly placed, non-overlapping distractor objects with varying shapes and colors. The dataset comprises 2,000 images, divided into perfectly balanced training and testing sets of 1,000 samples each. Examples of this benchmark are shown in Figure~\ref{fig:shapes_examples}.

\subsubsection{Common Perception–Reasoning Interface}
We decompose the system into:
(i) a neural perception unit producing distributions over object attributes and pairwise relations; and
(ii) a symbolic reasoning unit that consumes these as soft facts over the object domain and evaluates a single existential rule encoding the query. Supervision is a binary yes/no label optimized via a cross-entropy loss on the final answer distribution.

\subsubsection{DomiKnowS}
\textbf{Problem Specification.} DomiKnowS formulates the problem as a graph representation. It first declares the \textit{image} concept to represent the visual input. The \textit{objects} concept is then defined to represent the individual objects contained within the image. An explicit connection declaration is established in the framework between images and objects, indicating that each image may contain multiple objects. 
Under the image concept, two additional concepts—\textit{color} and \textit{shape}—are introduced to represent the attributes of each object for each shape and color considered. These two concepts serve as the outputs of the neural modeling component. Next, DomiKnowS defines a \textit{relation} concept between pairs of objects to capture their relational structure within the same image.
Finally, symbolic reasoning is expressed using \textit{existsL}, which denotes the existence of a particular combination of queries. The inference takes the following form:

{\centering
\texttt{existsL(is\_red(X), is\_blue(Y), is\_circle(Y), relation(Rel))}\par
}

where the $X$ and $Y$ represent the first object and second object, $Rel$ represent the relation of $X$ and $Y$. 

\noindent\textbf{Neural Modeling.} DomiKnowS uses a class of functions called \textit{ReaderSensor}s to process the raw input data. A \textit{ModuleLearner} is then employed to query the object-centric encoder, producing representations of the objects within the image. These representations are subsequently passed to three \textit{ModuleLearner} to predict the color (red and blue) and shape (circle) attributes of the objects. A \textit{CompositionCandidateSensor} is used to connect pairs of objects within the image, providing the information needed to compute their relations. Finally, the framework aggregates all information through predefined logical expressions, which are used to infer the final output.

\subsubsection{DeepProbLog}
\paragraph{Problem Specification.}
We use a ProbLog program with neural annotated disjunctions for
\texttt{color(Image,O,C)}, \texttt{shape(Image,O,S)}, and \texttt{relation(Image,O1,O2,R)}.
The decision rule is:

{\centering
\texttt{answer(Image, yes) :- obj(O1), obj(O2),}\\
\texttt{\ \ \ O1 /= O2,}\\
\texttt{\ \ \ color(Image,O1,red),}\\
\texttt{\ \ \ color(Image,O2,blue),}\\
\texttt{\ \ \ shape(Image,O2,circle),}\\
\texttt{\ \ \ relation(Image,O1,O2,R).}\par
}

A complementary \texttt{answer(Image, no)} ensures a normalized binary outcome. The object domain $\texttt{obj/1}$ is declared up to $n_{\max}$ derived from annotations. Batched queries are issued as \texttt{answer(tensor(batch(I)), Y)} and solved with an exact engine.

\paragraph{Neural Modeling.}
An object-centric encoder crops and embeds objects; heads output categorical distributions for color and shape per object and for relation per ordered object pair. These heads are wrapped as DeepProbLog networks and bound to the program’s neural predicates, enabling end-to-end training from the answer predicate.

\subsubsection{Scallop}
\paragraph{Problem Specification.}
We declare unary predicates over object indices for attributes (\texttt{red(i)}, \texttt{blue(j)}, \texttt{circle(j)}) and a binary predicate for the chosen spatial relation \texttt{R(i,j)}. The query is encoded as a Horn rule:

\[
\text{\texttt{ans()} :- \texttt{red(i)}, \texttt{blue(j)}, \texttt{circle(j)}, \texttt{above(i, j)}.}
\]

Scallop maps soft facts to weighted relations under a differentiable provenance and produces a normalized \{\texttt{yes},\texttt{no}\} answer.

\paragraph{Neural Modeling.}
Per-object (color, shape) and per-pair (relation) distributions from the encoder are converted to Scallop facts, respecting masks for padded objects and pairs. The Scallop context composes these facts with the rule to yield the answer distribution, providing gradients to the neural unit.

\subsubsection{LEFT}
\paragraph{Problem Specification.}
We express the query in first-order logic executed by LEFT’s generalized FOL executor:
\begin{center}
\texttt{exists(Object, lambda x: exists(Object, lambda y:}\\
\texttt{\ \ \ above(y, x) and red(x) and circle(y) and  blue(y)))}
\end{center}

The executor grounds over the object domain and evaluates the formula using attribute predicates for unary properties and a binary predicate for the relation, returning a scalar decision consistent with the query.

\paragraph{Neural Modeling.}
The object-centric encoder yields per-object attribute scores (for color/shape) and per-pair relation scores. These tensors parameterize the executor’s predicates, forming a differentiable perception–reasoning pipeline trained on the binary supervision.

\subsection{Toy NER}

For a simplified, toy version of the Named-Entity Recognition task~\citep{tjong-kim-sang-de-meulder-2003-introduction}, we create a dataset of randomly generated embeddings representing persons and locations. The objective is to learn the concepts of "works\_in", i.e., whether a person works in a location, and "is\_real\_person", i.e., whether a given embedding is a person or not. These concepts are learned using indirect supervision on two queries, that compose the atomic concepts.

{\centering
\texttt{constraint1 = is\_real\_person(P1) AND works\_in(P1,L1) AND is\_real\_person(P2) AND works\_in(P2,L2)}\\
\texttt{constraint2 = is\_real\_person(P2) AND works\_in(P2,L2) OR is\_real\_person(P3) AND works\_in(P3,L3)}\par
}

Here, $P1,P2,P3,L1,L2,L3$ are input embeddings with Ps representing persons and Ls representing locations. Thus, the final task is to predict the output of these two constraints (true or false), given 6 embeddings corresponding to 3 persons and 3 locations.

\subsubsection{DomiKnowS}
\textbf{Problem Specification.} 
To perform the Toy NER task, DomiKnowS begins by constructing a graph representation of the problem. It first defines the \textit{person} and \textit{location} concepts to represent the entity types of interest. Under the \textit{person} concept, three sub-concepts are declared to represent three individual persons, and under the location concept, three sub-concepts are introduced to represent distinct locations. A \textit{pair} relation is then defined to connect a person and a location. This relation is used to create three separate \textit{work\_in[i]} concepts, each representing the output relation between person $i$ and location $i$. Finally, the inference queries are expressed in a form-based manner, using \textit{andL} to represent logical conjunctions (AND) between concepts and \textit{orL} to represent logical disjunctions (OR) in the desired queries above.

\noindent\textbf{Neural Modeling.} The model declaration follows the standard pipeline of neural components. It begins with a \textit{ReaderSensor} to process the raw input data. A \textit{ModuleLearner} is then invoked, using the embeddings of either person or location to generate predictions based on the given representations. A \textit{JointSensor} is subsequently employed to connect the person and location concepts in order to form the \textit{work\_in[i]} relation specified in the problem definition. Finally, model inference is carried out based on predefined queries, yielding the final output.

\subsubsection{DeepProbLog}
\textbf{Problem Specification.} DeepProbLog declares 6 concepts, \textit{is\_real\_person[i]} for each person i and \textit{works\_in[i]} for each location i. These concepts are used to formulate the two constraints with the final query, \textit{check}, being the conjunction of the two. Note that, DeepProbLog does not support multiple simultaneous queries and hence, the two constraints cannot be trained with individual supervision for labels of each.

\noindent\textbf{Neural Modeling.} DeepProbLog uses the two neural networks for classification of the two concepts. The main challenge is the creation of the dataloader that reads from the raw data and converts it into the format necessary to send queries with the appropriate value substitutions for variables. This integration with ProbLog needs to be done by the user from scratch.

\subsubsection{Scallop}
\textbf{Problem Specification.} Scallop utilizes 6 concepts similar to DeepProbLog. The value of the two constraints is merged using a final module \textit{check(P1* P2 * W1 * W2 + P3 * W3)}. Here, the results of concepts \textit{is\_real\_person[i]} (P1, P2, P3) and \textit{works\_in[i]} (W1, W2, W3) are passed instead of the embeddings themselves. This context is passed through python itself, instead of a separate Datalog file.

\noindent\textbf{Neural Modeling.} The neural modeling component is standard with flexible loss declarations. To adapt from the raw data, an embedding generator \textit{NERDataset} is created which is the same as the \textit{JSONDataset} utilized in the DeepProbLog solution. However, there is no need to manually adapt the data to generate queries to the datalog component, as the forward function of the neural network is the final query and only requires the outputs of the neural networks passed to it.

\subsection{Math Equation Inference}

The Math Equation Inference task is designed to evaluate whether a neural network can learn local mathematical concepts from global supervision. 
The input to this task consists of two lists, each containing six real numbers sampled uniformly from the range $\left[-1,1\right]$. 
The objective is to determine whether the model can learn from a global condition that encompasses the properties of the first list, the properties of the second list, and the relationship between them. 
We consider two properties, that is, $\sum_{i=0}^8 x_i > 0 $ and $\sum_{i=0}^8 |x_i| > 0.5$. For relations between the two lists, we examine two cases: (i) whether the first elements of the lists have the same sign, and (ii) whether their last elements have opposite signs. 
These concepts are learned through indirect supervision using global queries that compose the atomic concepts, expressed as follows:

{\centering
\texttt{property1(L1) AND property2(L2) AND relation(L1, L2)}\par
}

where $L1, L2$ are randomly generated lists of six real numbers, property1 and property2 are drawn from the set of considered properties, and relation is drawn from the set of considered relations.

\subsubsection{DomiKnowS}\label{sec:DomiKnowS_math_ex}
\textbf{Problem Specification.} 
For implementing this task in DomiKnowS, the first concept defined is the \textit{problem} concept, which represents the overall mathematical inference problem. Next, the \textit{lst} concept is introduced to represent a list of considered numbers. Under this \textit{lst} concept, all possible condition concepts—\textit{is\_cond1} and \textit{is\_cond2}—are defined to represent the output class of each list. Subsequently, two relation concepts, \textit{is\_relation1} and \textit{is\_relation2}, are introduced between pairs of \textit{lst} instances to capture the two relational conditions described in the problem statement. Importantly, all possible conditions and relations must be defined, even if some are not ultimately used in the final inference.
To define the inference query, the logical operator andL is employed to combine the three relevant concepts. For example: \textit{andL(is\_cond1(L1), is\_relation1(L1, L2), is\_cond2(L2))}. This query corresponds to the case where the sum of $L1$ is greater than 0 (\textit{is\_cond1}), the first elements of $L1$ and $L2$ share the same sign (\textit{is\_relation1}), and the sum of the absolute values of $L2$ is greater than 0.5 (\textit{is\_cond2}).

\noindent\textbf{Neural Modeling.} 
DomiKnowS begins with a \textit{ReaderSensor} to process the two input lists of numbers. Then, \textit{CompositionCandidateSensor} is employed to connect the two lists, forming the intermediate representation required for relation prediction in later stages. Four neural networks are instantiated using \textit{ModuleLearner}—two dedicated to property concepts and two to relation concepts. 
These networks are trained to predict the respective properties and relations. The predicted concepts are subsequently integrated into the final query, which specifies the target relation for the given problem and serves as the prediction label during training and evaluation. It is important to note that DomiKnowS requires all possible inference queries to be explicitly represented in the graph. Each query is set as either active or inactive to obtain the correct prediction during inference.

\subsubsection{DeepProbLog}
\textbf{Problem Specification.} 
In this task, DeepProbLog defines four neural predicates: two for the considered properties (\textit{is\_property[i]\_nn} for object property i), and two for the considered relations (\textit{is\_relation[i]\_nn} for relation i).
These four neural predicates are then combined to define the final query concept, \textit{inference}. One \textit{inference} concept is created for each possible combination of the condition of L1 (2 possibilities), the condition of L2 (2 possibilities), and the relation between L1 and L2 (2 possibilities). In total, eight inference concepts are formulated. During evaluation, only the relevant concept corresponding to the input configuration is activated and used to produce the final prediction for the problem.

\noindent\textbf{Neural Modeling.} 
DeepProbLog employs four neural networks, corresponding to the defined concepts: two for property concepts and two for relation concepts. The pipeline then proceeds with data loading, where all possible candidate lists of numbers are read separately and pre-processed to obtain the required model inputs. 
This also includes manually connecting the list of numbers within the same problem. Then, all defined neural networks are called to obtain all possible relations and properties output.
Lastly, the query must be constructed in the exact predefined pattern to ensure that the system produces the correct output aligned with the desired inference condition.

\subsubsection{Scallop}
\textbf{Problem Specification.} 
Similar to DomiKnowS and DeepProbLog on this task, Scallop begins with the declaration of four concepts, that is, two property concepts and two relation concepts. However, Scallop only defines the one final module, \textit{inference(P1 * P2 * R)}, that accumulates the probability based on the output of the properties of $L1$ and $L2$, and the relation between $L1$ and $L2$. The output is defined in the neural modeling part to get the connection between the final module and the defined concept.

\noindent\textbf{Neural Modeling.} 
The neural modeling component in Scallop follows the standard pipeline. However, Scallop requires a manually defined connection between the model output and the declared final module. This requirement arises because the final module is specified in a general form, rather than being tied to a particular combination of inference concepts. In contrast, other frameworks explicitly define every possible query corresponding to a combination of properties and relations and refer only to those queries during inference.

\vspace{-2mm}
\section{Discussion and Future Direction}
 
Table \ref{tab:frameworks} summarizes the comparative aspects of existing frameworks and outlines future directions for optimizing, as we observe many columns marked with '\xmark', implying most frameworks present challenges that hinder the application and flexibility of the frameworks.  
While current frameworks are functional, future developments should take a more holistic approach that considers all aspects from an end-user perspective, aiming to improve usability as general-purpose libraries and foster wider adoption of neurosymbolic methods.

\noindent\textbf{Symbolic Representation.}
The generic neurosymbolic frameworks provide a formal knowledge representation language of their choice. 
The selected languages often are based on pure logical formalisms with established formal semantics, for example, Datalog or Prolog. However, we argue that knowledge representation for neurosymbolic frameworks  needs to be an innovative language designed for this integration purpose with adaptable semantics with learning as the pivotal concept~\citep{Kordjamshidi2019DeclarativeLP}. Restricting these frameworks to classical AI formalisms and formal semantics  limits the level of extension that can be made and restricts the support of various algorithms and types of integration.  

\noindent\s{\textbf{Neural Modeling.} Most of the examined frameworks leave neural modeling and the task of connecting the symbolic and sub-symbolic components, up to the user. This connection usually requires low-level data preprocessing, which is time consuming to implement. A lack of user-friendly libraries discourages developers from using neurosymbolic methods to solve downstream tasks. There is a need for abstractions in these frameworks~\citep{10.3389/frai.2022.755361} that improve user experience and remove the need for users to implement such data processing from scratch.}

\noindent\textbf{Model Declaration}. 
There is a need to be explicit about the low-level components of the neural architecture, enabling us to design interactions between neural and symbolic components and connect them as intended. The goal is to provide flexibility in designing arbitrary loss functions and connecting them to data for supervising concepts at various neural layers, which will allow any symbol to be learnable. 

\noindent\s{\textbf{Types of Interplay.}} Considering \cite{Kautz_2022}'s classification, current frameworks are limited in supporting one or two ways of interactions. The "Algo" column in Table \ref{tab:frameworks} shows that DeepProbLog and Scallop utilize one form of implementation, while DomiKnowS has multiple settings. 
One of the key challenges is determining the appropriate level of abstraction in a neural model after which reasoning should occur. The classification types demonstrate how a neural model can identify the relevant symbolic representations and suggest that neurosymbolic frameworks could leverage these models to learn and route inputs to the corresponding symbolic reasoning system. However, it remains unclear what level of abstraction is most effective for solving the end task in practice.

\noindent\s{\textbf{LLM.} Drawbacks often associated with employing symbolic AI into neural computing, such as creation of the symbolic knowledge for integration, can be mediated with the use of LLMs and foundation models.  LLMs have the potential to alleviate the classical issues in symbolic processing. Their vast knowledge can also be utilized to reduce the need for rebuilding neural components, allowing for flexible connections with different symbolic components.}

\vspace{-2mm}
\section{Conclusion}
Neurosymbolic AI presents a promising path forward in addressing the limitations of purely symbolic or neural approaches to AI. By integrating symbolic reasoning with neural learning, NeSy frameworks offer a balance between interpretability, data and time efficiency, and generalization. In this paper, we characterize the core components of NeSy frameworks and provide an analysis of some existing ones - DeepProbLog, Scallop, and DomiKnowS, illustrating the comparative facets. We identified some facets as symbolic knowledge and data representation, neural modeling, model declaration, method of integrating the symbolic and sub-symbolic systems, and role of LLMs. We identify key challenges in each facet that can guide us toward building the next generation of neurosymbolic frameworks. Unifying ideas in the field and building flexible frameworks by incorporating strengths in every facet will ease the learning curve associated with NeSy systems and improve standardization. Future NeSy frameworks should aim to provide flexible implementation, a user-friendly interface, improve scalability, and develop seamless integrations with foundation models. The advent of next-generation LLMs/VLMs provides promising solutions to longstanding knowledge engineering challenges, fostering more effective and scalable integration of symbolic representations and advancing research in neurosymbolic AI.

\acks{This project is partially supported by the Office of Naval Research (ONR) grant N00014-23-1-2417. Any opinions, findings, conclusions, or recommendations expressed in this material are those of the authors and do not necessarily reflect the views of the Office of Naval Research. We thank Uzair Mohammad for his editing and suggestions for Figure 1.}

\bibliography{references}

\begin{thebibliography}{91}
\providecommand{\natexlab}[1]{#1}
\providecommand{\url}[1]{\texttt{#1}}
\expandafter\ifx\csname urlstyle\endcsname\relax
  \providecommand{\doi}[1]{doi: #1}\else
  \providecommand{\doi}{doi: \begingroup \urlstyle{rm}\Url}\fi

\bibitem[Abiteboul et~al.(1995)Abiteboul, Hull, and Vianu]{10.5555/551350}
Serge Abiteboul, Richard Hull, and Victor Vianu.
\newblock \emph{Foundations of Databases: The Logical Level}.
\newblock Addison-Wesley Longman Publishing Co., Inc., USA, 1st edition, 1995.
\newblock ISBN 0201537710.

\bibitem[Acharya et~al.(2024)Acharya, Velasquez, and Song]{survey_archarya}
Kamal Acharya, Alvaro Velasquez, and Houbing~Herbert Song.
\newblock A survey on symbolic knowledge distillation of large language models.
\newblock \emph{IEEE Transactions on Artificial Intelligence}, 5\penalty0
  (12):\penalty0 5928--5948, 2024.
\newblock \doi{10.1109/TAI.2024.3428519}.

\bibitem[Aditya et~al.(2023)Aditya, Mukherji, Balasubramanian, Chaudhary, and
  Shakarian]{aditya2023pyreasonsoftwareopenworld}
Dyuman Aditya, Kaustuv Mukherji, Srikar Balasubramanian, Abhiraj Chaudhary, and
  Paulo Shakarian.
\newblock Pyreason: Software for open world temporal logic, 2023.
\newblock URL \url{https://arxiv.org/abs/2302.13482}.

\bibitem[Ahmad et~al.(2019)Ahmad, Farman, and Jan]{Ahmad2019}
Jamil Ahmad, Haleem Farman, and Zahoor Jan.
\newblock \emph{Deep Learning Methods and Applications}, pages 31--42.
\newblock Springer Singapore, Singapore, 2019.
\newblock ISBN 978-981-13-3459-7.
\newblock \doi{10.1007/978-981-13-3459-7_3}.
\newblock URL \url{https://doi.org/10.1007/978-981-13-3459-7_3}.

\bibitem[Ahmed et~al.(2022)Ahmed, Li, Ton, Guo, Chang, Kordjamshidi, Srikumar,
  Van~den Broeck, and Singh]{pmlr-v176-ahmed22a}
Kareem Ahmed, Tao Li, Thy Ton, Quan Guo, Kai-Wei Chang, Parisa Kordjamshidi,
  Vivek Srikumar, Guy Van~den Broeck, and Sameer Singh.
\newblock Pylon: A pytorch framework for learning with constraints.
\newblock In Douwe Kiela, Marco Ciccone, and Barbara Caputo, editors,
  \emph{Proceedings of the NeurIPS 2021 Competitions and Demonstrations Track},
  volume 176 of \emph{Proceedings of Machine Learning Research}, pages
  319--324. PMLR, 06--14 Dec 2022.
\newblock URL \url{https://proceedings.mlr.press/v176/ahmed22a.html}.

\bibitem[Augusto(2021)]{Augusto2021-AUGFST-2}
Luis~M. Augusto.
\newblock From symbols to knowledge systems: A. newell and h. a. {S}imon's
  contribution to symbolic {AI}.
\newblock \emph{Journal of Knowledge Structures and Systems}, 2\penalty0
  (1):\penalty0 29--62, 2021.

\bibitem[Badreddine et~al.(2022)Badreddine, {d'Avila Garcez}, Serafini, and
  Spranger]{BADREDDINE2022103649}
Samy Badreddine, Artur {d'Avila Garcez}, Luciano Serafini, and Michael
  Spranger.
\newblock Logic tensor networks.
\newblock \emph{Artificial Intelligence}, 303:\penalty0 103649, 2022.
\newblock ISSN 0004-3702.
\newblock \doi{https://doi.org/10.1016/j.artint.2021.103649}.
\newblock URL
  \url{https://www.sciencedirect.com/science/article/pii/S0004370221002009}.

\bibitem[Bender et~al.(2021)Bender, Gebru, McMillan-Major, and
  Shmitchell]{10.1145/3442188.3445922}
Emily~M. Bender, Timnit Gebru, Angelina McMillan-Major, and Shmargaret
  Shmitchell.
\newblock On the dangers of stochastic parrots: Can language models be too big?
\newblock In \emph{Proceedings of the 2021 ACM Conference on Fairness,
  Accountability, and Transparency}, FAccT '21, page 610–623, New York, NY,
  USA, 2021. Association for Computing Machinery.
\newblock ISBN 9781450383097.
\newblock \doi{10.1145/3442188.3445922}.
\newblock URL \url{https://doi.org/10.1145/3442188.3445922}.

\bibitem[Bhuyan et~al.(2024)Bhuyan, Ramdane-Cherif, Tomar, and
  Singh]{Bhuyan2024}
Bikram~Pratim Bhuyan, Amar Ramdane-Cherif, Ravi Tomar, and T.~P. Singh.
\newblock Neuro-symbolic artificial intelligence: a survey.
\newblock \emph{Neural Computing and Applications}, 36\penalty0 (21):\penalty0
  12809--12844, July 2024.
\newblock ISSN 1433-3058.
\newblock \doi{10.1007/s00521-024-09960-z}.
\newblock URL \url{https://doi.org/10.1007/s00521-024-09960-z}.

\bibitem[Booch et~al.(2021)Booch, Fabiano, Horesh, Kate, Lenchner, Linck,
  Loreggia, Murgesan, Mattei, Rossi, and
  Srivastava]{Booch_Fabiano_Horesh_Kate_Lenchner_Linck_Loreggia_Murgesan_Mattei_Rossi_Srivastava_2021}
Grady Booch, Francesco Fabiano, Lior Horesh, Kiran Kate, Jonathan Lenchner,
  Nick Linck, Andreas Loreggia, Keerthiram Murgesan, Nicholas Mattei, Francesca
  Rossi, and Biplav Srivastava.
\newblock Thinking fast and slow in ai.
\newblock \emph{Proceedings of the AAAI Conference on Artificial Intelligence},
  35\penalty0 (17):\penalty0 15042--15046, May 2021.
\newblock \doi{10.1609/aaai.v35i17.17765}.
\newblock URL \url{https://ojs.aaai.org/index.php/AAAI/article/view/17765}.

\bibitem[Bouneffouf and
  Aggarwal(2022)]{bouneffouf2022surveyapplicationsneurosymbolicartificial}
Djallel Bouneffouf and Charu~C. Aggarwal.
\newblock Survey on applications of neurosymbolic artificial intelligence,
  2022.
\newblock URL \url{https://arxiv.org/abs/2209.12618}.

\bibitem[Bratko and Muggleton(1995)]{10.1145/219717.219771}
Ivan Bratko and Stephen Muggleton.
\newblock Applications of inductive logic programming.
\newblock \emph{Commun. ACM}, 38\penalty0 (11):\penalty0 65–70, November
  1995.
\newblock ISSN 0001-0782.
\newblock \doi{10.1145/219717.219771}.
\newblock URL \url{https://doi.org/10.1145/219717.219771}.

\bibitem[Burattini et~al.(2002)Burattini, de~Francesco, and
  De~Gregorio]{1181487}
E.~Burattini, A.~de~Francesco, and M.~De~Gregorio.
\newblock Nsl: a neuro-symbolic language for monotonic and non-monotonic
  logical inferences.
\newblock In \emph{VII Brazilian Symposium on Neural Networks, 2002. SBRN 2002.
  Proceedings.}, pages 256--261, 2002.
\newblock \doi{10.1109/SBRN.2002.1181487}.

\bibitem[Clocksin and Mellish(2003)]{clocksin2003programming}
William~F Clocksin and Christopher~S Mellish.
\newblock \emph{Programming in PROLOG}.
\newblock Springer Science \& Business Media, 2003.

\bibitem[Cohen et~al.(2017)Cohen, Yang, and Mazaitis]{cohen2017tensorlog}
William~W Cohen, Fan Yang, and Kathryn~Rivard Mazaitis.
\newblock Tensorlog: Deep learning meets probabilistic dbs.
\newblock \emph{arXiv preprint arXiv:1707.05390}, 2017.

\bibitem[Cropper and Duman\v{c}i\'{c}(2022)]{10.1613/jair.1.13507}
Andrew Cropper and Sebastijan Duman\v{c}i\'{c}.
\newblock Inductive logic programming at 30: A new introduction.
\newblock \emph{J. Artif. Int. Res.}, 74, September 2022.
\newblock ISSN 1076-9757.
\newblock \doi{10.1613/jair.1.13507}.
\newblock URL \url{https://doi.org/10.1613/jair.1.13507}.

\bibitem[Darwiche(2011)]{darwiche2011sdd}
Adnan Darwiche.
\newblock Sdd: A new canonical representation of propositional knowledge bases.
\newblock In \emph{IJCAI Proceedings-International Joint Conference on
  Artificial Intelligence}, volume~22, page 819, 2011.

\bibitem[De~Raedt et~al.(2007)De~Raedt, Kimmig, and
  Toivonen]{10.5555/1625275.1625673}
Luc De~Raedt, Angelika Kimmig, and Hannu Toivonen.
\newblock Problog: a probabilistic prolog and its application in link
  discovery.
\newblock In \emph{Proceedings of the 20th International Joint Conference on
  Artifical Intelligence}, IJCAI'07, page 2468–2473, San Francisco, CA, USA,
  2007. Morgan Kaufmann Publishers Inc.

\bibitem[Derkinderen et~al.(2025)Derkinderen, Manhaeve, Adriaensen, Praet,
  Smet, Marra, and Raedt]{derkinderen2025deeplogneurosymbolicmachine}
Vincent Derkinderen, Robin Manhaeve, Rik Adriaensen, Lucas~Van Praet,
  Lennert~De Smet, Giuseppe Marra, and Luc~De Raedt.
\newblock The deeplog neurosymbolic machine, 2025.
\newblock URL \url{https://arxiv.org/abs/2508.13697}.

\bibitem[Eisner(2002)]{eisner2002parameter}
Jason Eisner.
\newblock Parameter estimation for probabilistic finite-state transducers.
\newblock In \emph{Proceedings of the 40th Annual Meeting of the Association
  for Computational Linguistics}, pages 1--8, 2002.

\bibitem[Fabiano et~al.(2023)Fabiano, Pallagani, Ganapini, Horesh, Loreggia,
  Murugesan, Rossi, and Srivastava]{fabiano2023plansofai}
Francesco Fabiano, Vishal Pallagani, Marianna~Bergamaschi Ganapini, Lior
  Horesh, Andrea Loreggia, Keerthiram Murugesan, Francesca Rossi, and Biplav
  Srivastava.
\newblock Plan-{SOFAI}: A neuro-symbolic planning architecture.
\newblock In \emph{Neuro-Symbolic Learning and Reasoning in the era of Large
  Language Models}, 2023.
\newblock URL \url{https://openreview.net/forum?id=ORAhay0H4x}.

\bibitem[Faghihi et~al.(2023)Faghihi, Nafar, Zheng, Mirzaee, Zhang, Uszok, Wan,
  Premsri, Roth, and Kordjamshidi]{faghihi2023glueconsgenericbenchmarklearning}
Hossein~Rajaby Faghihi, Aliakbar Nafar, Chen Zheng, Roshanak Mirzaee, Yue
  Zhang, Andrzej Uszok, Alexander Wan, Tanawan Premsri, Dan Roth, and Parisa
  Kordjamshidi.
\newblock Gluecons: A generic benchmark for learning under constraints, 2023.
\newblock URL \url{https://arxiv.org/abs/2302.10914}.

\bibitem[Faghihi et~al.(2024)Faghihi, Nafar, Uszok, Karimian, and
  Kordjamshidi]{faghihi2024prompt2demodeldeclarativeneurosymbolicmodeling}
Hossein~Rajaby Faghihi, Aliakbar Nafar, Andrzej Uszok, Hamid Karimian, and
  Parisa Kordjamshidi.
\newblock Prompt2demodel: Declarative neuro-symbolic modeling with natural
  language, 2024.
\newblock URL \url{https://arxiv.org/abs/2407.20513}.

\bibitem[Fang and Yu(2024)]{fang2024large}
Chuyu Fang and Song-Chun Yu.
\newblock Large language models are neurosymbolic reasoners.
\newblock \emph{arXiv preprint arXiv:2401.09334}, 2024.
\newblock URL \url{https://arxiv.org/html/2401.09334v1}.

\bibitem[Frazier and Pitt(1993)]{frazier1993learning}
Michael Frazier and Leonard Pitt.
\newblock Learning from entailment: An application to propositional horn
  sentences.
\newblock In \emph{Proceedings of the Tenth International Conference on
  International Conference on Machine Learning}, pages 120--127, 1993.

\bibitem[Giunchiglia et~al.(2024)Giunchiglia, Tatomir, Stoian, and
  Lukasiewicz]{GIUNCHIGLIA2024109124}
Eleonora Giunchiglia, Alex Tatomir, Mihaela~Cătălina Stoian, and Thomas
  Lukasiewicz.
\newblock Ccn+: A neuro-symbolic framework for deep learning with requirements.
\newblock \emph{International Journal of Approximate Reasoning}, 171:\penalty0
  109124, 2024.
\newblock ISSN 0888-613X.
\newblock \doi{https://doi.org/10.1016/j.ijar.2024.109124}.
\newblock URL
  \url{https://www.sciencedirect.com/science/article/pii/S0888613X24000112}.
\newblock Synergies between Machine Learning and Reasoning.

\bibitem[Green et~al.(2007)Green, Karvounarakis, and
  Tannen]{10.1145/1265530.1265535}
Todd~J. Green, Grigoris Karvounarakis, and Val Tannen.
\newblock Provenance semirings.
\newblock In \emph{Proceedings of the Twenty-Sixth ACM SIGMOD-SIGACT-SIGART
  Symposium on Principles of Database Systems}, PODS '07, page 31–40, New
  York, NY, USA, 2007. Association for Computing Machinery.
\newblock ISBN 9781595936851.
\newblock \doi{10.1145/1265530.1265535}.
\newblock URL \url{https://doi.org/10.1145/1265530.1265535}.

\bibitem[Guo et~al.(2020)Guo, Rajaby~Faghihi, Zhang, Uszok, and
  Kordjamshidi]{ijcai2020p382}
Quan Guo, Hossein Rajaby~Faghihi, Yue Zhang, Andrzej Uszok, and Parisa
  Kordjamshidi.
\newblock Inference-masked loss for deep structured output learning.
\newblock In Christian Bessiere, editor, \emph{Proceedings of the Twenty-Ninth
  International Joint Conference on Artificial Intelligence, {IJCAI-20}}, pages
  2754--2761. International Joint Conferences on Artificial Intelligence
  Organization, 7 2020.
\newblock \doi{10.24963/ijcai.2020/382}.
\newblock URL \url{https://doi.org/10.24963/ijcai.2020/382}.
\newblock Main track.

\bibitem[{Gurobi Optimization, LLC}(2024)]{gurobi}
{Gurobi Optimization, LLC}.
\newblock {Gurobi Optimizer Reference Manual}, 2024.
\newblock URL \url{https://www.gurobi.com}.

\bibitem[Hayes-Roth(1985)]{hayes1985rule}
Frederick Hayes-Roth.
\newblock Rule-based systems.
\newblock \emph{Communications of the ACM}, 28\penalty0 (9):\penalty0 921--932,
  1985.

\bibitem[Hitzler and Sarker(2022)]{hitzler2022neuro}
Pascal Hitzler and Md~Kamruzzaman Sarker.
\newblock Neuro-symbolic artificial intelligence: The state of the art.
\newblock 2022.

\bibitem[Hossain and
  Chen(2025)]{hossain2025studyneurosymbolicartificialintelligence}
Delower Hossain and Jake~Y Chen.
\newblock A study on neuro-symbolic artificial intelligence: Healthcare
  perspectives, 2025.
\newblock URL \url{https://arxiv.org/abs/2503.18213}.

\bibitem[Hsu et~al.(2023)Hsu, Mao, Tenenbaum, and
  Wu]{hsu2023whatsleftconceptgrounding}
Joy Hsu, Jiayuan Mao, Joshua~B. Tenenbaum, and Jiajun Wu.
\newblock What's left? concept grounding with logic-enhanced foundation models,
  2023.
\newblock URL \url{https://arxiv.org/abs/2310.16035}.

\bibitem[Huang et~al.(2021)Huang, Li, Chen, Samel, Naik, Song, and
  Si]{NEURIPS2021_d367eef1}
Jiani Huang, Ziyang Li, Binghong Chen, Karan Samel, Mayur Naik, Le~Song, and
  Xujie Si.
\newblock Scallop: From probabilistic deductive databases to scalable
  differentiable reasoning.
\newblock In M.~Ranzato, A.~Beygelzimer, Y.~Dauphin, P.S. Liang, and J.~Wortman
  Vaughan, editors, \emph{Advances in Neural Information Processing Systems},
  volume~34, pages 25134--25145. Curran Associates, Inc., 2021.
\newblock URL
  \url{https://proceedings.neurips.cc/paper_files/paper/2021/file/d367eef13f90793bd8121e2f675f0dc2-Paper.pdf}.

\bibitem[Ishay et~al.(2023)Ishay, Yang, and Lee]{Ishay2023LLM2ASP}
Adam Ishay, Zhun Yang, and Joohyung Lee.
\newblock Leveraging large language models to generate answer set programs.
\newblock In Pierre Marquis, Tran~Cao Son, and Gabriele Kern-Isberner, editors,
  \emph{Proceedings of the 20th International Conference on Principles of
  Knowledge Representation and Reasoning, KR 2023}, Proceedings of the
  International Conference on Knowledge Representation and Reasoning, pages
  374--383. Association for the Advancement of Artificial Intelligence, 2023.
\newblock \doi{10.24963/kr.2023/37}.

\bibitem[Jayasingha et~al.(2025)Jayasingha, Iancu, and Lilius]{10981497}
Prashani Jayasingha, Bogdan Iancu, and Johan Lilius.
\newblock Neurosymbolic approaches in ai design – an overview.
\newblock In \emph{2025 IEEE Symposium on Trustworthy, Explainable and
  Responsible Computational Intelligence (CITREx Companion)}, pages 1--5, 2025.
\newblock \doi{10.1109/CITRExCompanion65208.2025.10981497}.

\bibitem[Johnson et~al.(2017)Johnson, Hariharan, van~der Maaten, Fei-Fei,
  Lawrence~Zitnick, and Girshick]{Johnson_2017_CVPR}
Justin Johnson, Bharath Hariharan, Laurens van~der Maaten, Li~Fei-Fei,
  C.~Lawrence~Zitnick, and Ross Girshick.
\newblock Clevr: A diagnostic dataset for compositional language and elementary
  visual reasoning.
\newblock In \emph{Proceedings of the IEEE Conference on Computer Vision and
  Pattern Recognition (CVPR)}, July 2017.

\bibitem[Kahneman(2011)]{kahneman2011thinking}
Daniel Kahneman.
\newblock \emph{Thinking, fast and slow}.
\newblock macmillan, 2011.

\bibitem[Kamali et~al.(2025)Kamali, Barezi, and Kordjamshidi]{nesycoco}
Danial Kamali, Elham~J. Barezi, and Parisa Kordjamshidi.
\newblock Nesycoco: A neuro-symbolic concept composer for compositional
  generalization.
\newblock \emph{Proceedings of the AAAI Conference on Artificial Intelligence},
  39\penalty0 (4):\penalty0 4184--4193, Apr. 2025.
\newblock \doi{10.1609/aaai.v39i4.32439}.
\newblock URL \url{https://ojs.aaai.org/index.php/AAAI/article/view/32439}.

\bibitem[Kautz(2022)]{Kautz_2022}
Henry Kautz.
\newblock The third ai summer: Aaai robert s. engelmore memorial lecture.
\newblock \emph{AI Magazine}, 43\penalty0 (1):\penalty0 105--125, Mar. 2022.
\newblock \doi{10.1002/aaai.12036}.
\newblock URL
  \url{https://ojs.aaai.org/aimagazine/index.php/aimagazine/article/view/19122}.

\bibitem[Kimmig et~al.(2011)Kimmig, Van~den Broeck, and
  De~Raedt]{Kimmig_VandenBroeck_DeRaedt_2011}
Angelika Kimmig, Guy Van~den Broeck, and Luc De~Raedt.
\newblock An algebraic prolog for reasoning about possible worlds.
\newblock \emph{Proceedings of the AAAI Conference on Artificial Intelligence},
  25\penalty0 (1):\penalty0 209--214, Aug. 2011.
\newblock \doi{10.1609/aaai.v25i1.7852}.
\newblock URL \url{https://ojs.aaai.org/index.php/AAAI/article/view/7852}.

\bibitem[Kirillov et~al.(2023)Kirillov, Mintun, Ravi, Mao, Rolland, Gustafson,
  Xiao, Whitehead, Berg, Lo, Dollár, and Girshick]{kirillov2023segment}
Alexander Kirillov, Eric Mintun, Nikhila Ravi, Hanzi Mao, Chloe Rolland, Laura
  Gustafson, Tete Xiao, Spencer Whitehead, Alexander~C. Berg, Wan-Yen Lo, Piotr
  Dollár, and Ross Girshick.
\newblock Segment anything, 2023.
\newblock URL \url{https://arxiv.org/abs/2304.02643}.

\bibitem[Kolaitis and Vardi(1990)]{10.1145/298514.298542}
Phokion~G. Kolaitis and Moshe~Y. Vardi.
\newblock On the expressive power of datalog: tools and a case study.
\newblock In \emph{Proceedings of the Ninth ACM SIGACT-SIGMOD-SIGART Symposium
  on Principles of Database Systems}, PODS '90, page 61–71, New York, NY,
  USA, 1990. Association for Computing Machinery.
\newblock ISBN 0897913523.
\newblock \doi{10.1145/298514.298542}.
\newblock URL \url{https://doi.org/10.1145/298514.298542}.

\bibitem[Kordjamshidi et~al.(2015)Kordjamshidi, Roth, and Wu]{kjam}
Parisa Kordjamshidi, Dan Roth, and Hao Wu.
\newblock Saul: Towards declarative learning based programming.
\newblock volume 2015, 12 2015.

\bibitem[Kordjamshidi et~al.(2016)Kordjamshidi, Khashabi, Christodoulopoulos,
  Mangipudi, Singh, and Roth]{kordjamshidi-etal-2016-better}
Parisa Kordjamshidi, Daniel Khashabi, Christos Christodoulopoulos, Bhargav
  Mangipudi, Sameer Singh, and Dan Roth.
\newblock Better call {S}aul: Flexible programming for learning and inference
  in {NLP}.
\newblock In Yuji Matsumoto and Rashmi Prasad, editors, \emph{Proceedings of
  {COLING} 2016, the 26th International Conference on Computational
  Linguistics: Technical Papers}, pages 3030--3040, Osaka, Japan, December
  2016. The COLING 2016 Organizing Committee.
\newblock URL \url{https://aclanthology.org/C16-1285/}.

\bibitem[Kordjamshidi et~al.(2019)Kordjamshidi, Roth, and
  Kersting]{Kordjamshidi2019DeclarativeLP}
Parisa Kordjamshidi, Dan Roth, and Kristian Kersting.
\newblock Declarative learning-based programming as an interface to ai systems.
\newblock \emph{Frontiers in Artificial Intelligence}, 5, 2019.
\newblock URL \url{https://api.semanticscholar.org/CorpusID:195069123}.

\bibitem[Kordjamshidi et~al.(2022)Kordjamshidi, Roth, and
  Kersting]{10.3389/frai.2022.755361}
Parisa Kordjamshidi, Dan Roth, and Kristian Kersting.
\newblock Declarative learning-based programming as an interface to ai systems.
\newblock \emph{Frontiers in Artificial Intelligence}, Volume 5 - 2022, 2022.
\newblock ISSN 2624-8212.
\newblock \doi{10.3389/frai.2022.755361}.
\newblock URL
  \url{https://www.frontiersin.org/journals/artificial-intelligence/articles/10.3389/frai.2022.755361}.

\bibitem[Lamb et~al.(2021)Lamb, Garcez, Gori, Prates, Avelar, and
  Vardi]{lamb2021graphneuralnetworksmeet}
Luis~C. Lamb, Artur Garcez, Marco Gori, Marcelo Prates, Pedro Avelar, and Moshe
  Vardi.
\newblock Graph neural networks meet neural-symbolic computing: A survey and
  perspective, 2021.
\newblock URL \url{https://arxiv.org/abs/2003.00330}.

\bibitem[Lample and Charton(2020)]{Lample2020Deep}
Guillaume Lample and François Charton.
\newblock Deep learning for symbolic mathematics.
\newblock In \emph{International Conference on Learning Representations}, 2020.
\newblock URL \url{https://openreview.net/forum?id=S1eZYeHFDS}.

\bibitem[Lecun et~al.(1998)Lecun, Bottou, Bengio, and Haffner]{726791}
Y.~Lecun, L.~Bottou, Y.~Bengio, and P.~Haffner.
\newblock Gradient-based learning applied to document recognition.
\newblock \emph{Proceedings of the IEEE}, 86\penalty0 (11):\penalty0
  2278--2324, 1998.
\newblock \doi{10.1109/5.726791}.

\bibitem[LeCun et~al.(2015)LeCun, Bengio, and Hinton]{DL}
Yann LeCun, Yoshua Bengio, and Geoffrey Hinton.
\newblock Deep learning.
\newblock \emph{Nature}, 521\penalty0 (7553):\penalty0 436--444, 2015.
\newblock \doi{10.1038/nature14539}.
\newblock URL \url{https://doi.org/10.1038/nature14539}.

\bibitem[Li et~al.(2023{\natexlab{a}})Li, Qi, Liu, Di, Liu, Pei, Yi, and
  Zhou]{10.1145/3555803}
Bo~Li, Peng Qi, Bo~Liu, Shuai Di, Jingen Liu, Jiquan Pei, Jinfeng Yi, and Bowen
  Zhou.
\newblock Trustworthy ai: From principles to practices.
\newblock \emph{ACM Comput. Surv.}, 55\penalty0 (9), January
  2023{\natexlab{a}}.
\newblock ISSN 0360-0300.
\newblock \doi{10.1145/3555803}.
\newblock URL \url{https://doi.org/10.1145/3555803}.

\bibitem[Li et~al.(2023{\natexlab{b}})Li, Huang, and Naik]{10.1145/3591280}
Ziyang Li, Jiani Huang, and Mayur Naik.
\newblock Scallop: A language for neurosymbolic programming.
\newblock \emph{Proc. ACM Program. Lang.}, 7\penalty0 (PLDI), June
  2023{\natexlab{b}}.
\newblock \doi{10.1145/3591280}.
\newblock URL \url{https://doi.org/10.1145/3591280}.

\bibitem[Li et~al.(2024)Li, Huang, Liu, Zhu, Zhao, Dodds, Velingker, Alur, and
  Naik]{Li_2024}
Ziyang Li, Jiani Huang, Jason Liu, Felix Zhu, Eric Zhao, William Dodds, Neelay
  Velingker, Rajeev Alur, and Mayur Naik.
\newblock Relational programming with foundational models.
\newblock \emph{Proceedings of the AAAI Conference on Artificial Intelligence},
  38\penalty0 (9):\penalty0 10635–10644, March 2024.
\newblock ISSN 2159-5399.
\newblock \doi{10.1609/aaai.v38i9.28934}.
\newblock URL \url{http://dx.doi.org/10.1609/aaai.v38i9.28934}.

\bibitem[Lima et~al.(2005)Lima, Morveli-Espinoza, Pereira, and
  França]{inproceedings}
Priscila Lima, Mariela Morveli-Espinoza, Glaucia K~E Pereira, and Felipe
  França.
\newblock Satyrus: A sat-based neuro-symbolic architecture for constraint
  processing.
\newblock pages 137--142, 01 2005.
\newblock \doi{10.1109/ICHIS.2005.97}.

\bibitem[Liu et~al.(2019)Liu, Liu, Bai, and Yuille]{Liu_2019_CVPR}
Runtao Liu, Chenxi Liu, Yutong Bai, and Alan~L. Yuille.
\newblock Clevr-ref+: Diagnosing visual reasoning with referring expressions.
\newblock In \emph{Proceedings of the IEEE/CVF Conference on Computer Vision
  and Pattern Recognition (CVPR)}, June 2019.

\bibitem[Lu et~al.(2024)Lu, Afridi, Kang, Ruchkin, and Zheng]{rel}
Zhen Lu, Imran Afridi, Hong~Jin Kang, Ivan Ruchkin, and Xi~Zheng.
\newblock Surveying neuro-symbolic approaches for reliable artificial
  intelligence of things.
\newblock \emph{Journal of Reliable Intelligent Environments}, 10\penalty0
  (3):\penalty0 257--279, 2024.
\newblock \doi{10.1007/s40860-024-00231-1}.
\newblock URL \url{https://doi.org/10.1007/s40860-024-00231-1}.

\bibitem[Manhaeve et~al.(2021)Manhaeve, Dumančić, Kimmig, Demeester, and {De
  Raedt}]{MANHAEVE2021103504}
Robin Manhaeve, Sebastijan Dumančić, Angelika Kimmig, Thomas Demeester, and
  Luc {De Raedt}.
\newblock Neural probabilistic logic programming in deepproblog.
\newblock \emph{Artificial Intelligence}, 298:\penalty0 103504, 2021.
\newblock ISSN 0004-3702.
\newblock \doi{https://doi.org/10.1016/j.artint.2021.103504}.
\newblock URL
  \url{https://www.sciencedirect.com/science/article/pii/S0004370221000552}.

\bibitem[Mao et~al.(2019)Mao, Gan, Kohli, Tenenbaum, and
  Wu]{mao2019neurosymbolicconceptlearnerinterpreting}
Jiayuan Mao, Chuang Gan, Pushmeet Kohli, Joshua~B. Tenenbaum, and Jiajun Wu.
\newblock The neuro-symbolic concept learner: Interpreting scenes, words, and
  sentences from natural supervision, 2019.
\newblock URL \url{https://arxiv.org/abs/1904.12584}.

\bibitem[Minderer et~al.(2022)Minderer, Gritsenko, Stone, Neumann, Weissenborn,
  Dosovitskiy, Mahendran, Arnab, Dehghani, Shen, Wang, Zhai, Kipf, and
  Houlsby]{minderer2022simpleopenvocabularyobjectdetection}
Matthias Minderer, Alexey Gritsenko, Austin Stone, Maxim Neumann, Dirk
  Weissenborn, Alexey Dosovitskiy, Aravindh Mahendran, Anurag Arnab, Mostafa
  Dehghani, Zhuoran Shen, Xiao Wang, Xiaohua Zhai, Thomas Kipf, and Neil
  Houlsby.
\newblock Simple open-vocabulary object detection with vision transformers,
  2022.
\newblock URL \url{https://arxiv.org/abs/2205.06230}.

\bibitem[Mirzaee and
  Kordjamshidi(2023)]{mirzaee-kordjamshidi-2023-disentangling}
Roshanak Mirzaee and Parisa Kordjamshidi.
\newblock Disentangling extraction and reasoning in multi-hop spatial
  reasoning.
\newblock In Houda Bouamor, Juan Pino, and Kalika Bali, editors, \emph{Findings
  of the Association for Computational Linguistics: EMNLP 2023}, pages
  3379--3397, Singapore, December 2023. Association for Computational
  Linguistics.
\newblock \doi{10.18653/v1/2023.findings-emnlp.221}.
\newblock URL \url{https://aclanthology.org/2023.findings-emnlp.221/}.

\bibitem[Nandwani et~al.(2019)Nandwani, Pathak, Mausam, and
  Singla]{NEURIPS2019_cf708fc1}
Yatin Nandwani, Abhishek Pathak, Mausam, and Parag Singla.
\newblock A primal dual formulation for deep learning with constraints.
\newblock In H.~Wallach, H.~Larochelle, A.~Beygelzimer, F.~d\textquotesingle
  Alch\'{e}-Buc, E.~Fox, and R.~Garnett, editors, \emph{Advances in Neural
  Information Processing Systems}, volume~32. Curran Associates, Inc., 2019.
\newblock URL
  \url{https://proceedings.neurips.cc/paper_files/paper/2019/file/cf708fc1decf0337aded484f8f4519ae-Paper.pdf}.

\bibitem[Newell(1980)]{NEWELL1980135}
Allen Newell.
\newblock Physical symbol systems.
\newblock \emph{Cognitive Science}, 4\penalty0 (2):\penalty0 135--183, 1980.
\newblock ISSN 0364-0213.
\newblock \doi{https://doi.org/10.1016/S0364-0213(80)80015-2}.
\newblock URL
  \url{https://www.sciencedirect.com/science/article/pii/S0364021380800152}.

\bibitem[Ng and Subrahmanian(1992)]{NG1992150}
Raymond Ng and V.S. Subrahmanian.
\newblock Probabilistic logic programming.
\newblock \emph{Information and Computation}, 101\penalty0 (2):\penalty0
  150--201, 1992.
\newblock ISSN 0890-5401.
\newblock \doi{https://doi.org/10.1016/0890-5401(92)90061-J}.
\newblock URL
  \url{https://www.sciencedirect.com/science/article/pii/089054019290061J}.

\bibitem[Nienhuys-Cheng and de~Wolf(1997)]{nienhuys1997inductive}
Shan-Hwei Nienhuys-Cheng and Roland de~Wolf.
\newblock \emph{What is inductive logic programming?}
\newblock Springer, 1997.

\bibitem[OpenAI et~al.(2024)OpenAI, Achiam, and
  et~al.]{openai2024gpt4technicalreport}
OpenAI, Josh Achiam, and Steven~Adler et~al.
\newblock Gpt-4 technical report, 2024.
\newblock URL \url{https://arxiv.org/abs/2303.08774}.

\bibitem[Pan et~al.(2023{\natexlab{a}})Pan, Albalak, Wang, and
  Wang]{pan-etal-2023-logic}
Liangming Pan, Alon Albalak, Xinyi Wang, and William Wang.
\newblock Logic-{LM}: Empowering large language models with symbolic solvers
  for faithful logical reasoning.
\newblock In Houda Bouamor, Juan Pino, and Kalika Bali, editors, \emph{Findings
  of the Association for Computational Linguistics: EMNLP 2023}, pages
  3806--3824, Singapore, December 2023{\natexlab{a}}. Association for
  Computational Linguistics.
\newblock \doi{10.18653/v1/2023.findings-emnlp.248}.
\newblock URL \url{https://aclanthology.org/2023.findings-emnlp.248/}.

\bibitem[Pan et~al.(2023{\natexlab{b}})Pan, Albalak, Wang, and
  Wang]{pan2023logiclmempoweringlargelanguage}
Liangming Pan, Alon Albalak, Xinyi Wang, and William~Yang Wang.
\newblock Logic-lm: Empowering large language models with symbolic solvers for
  faithful logical reasoning, 2023{\natexlab{b}}.
\newblock URL \url{https://arxiv.org/abs/2305.12295}.

\bibitem[Paszke et~al.(2019)Paszke, Gross, Massa, Lerer, Bradbury, Chanan,
  Killeen, Lin, Gimelshein, Antiga, Desmaison, Köpf, Yang, DeVito, Raison,
  Tejani, Chilamkurthy, Steiner, Fang, Bai, and
  Chintala]{paszke2019pytorchimperativestylehighperformance}
Adam Paszke, Sam Gross, Francisco Massa, Adam Lerer, James Bradbury, Gregory
  Chanan, Trevor Killeen, Zeming Lin, Natalia Gimelshein, Luca Antiga, Alban
  Desmaison, Andreas Köpf, Edward Yang, Zach DeVito, Martin Raison, Alykhan
  Tejani, Sasank Chilamkurthy, Benoit Steiner, Lu~Fang, Junjie Bai, and Soumith
  Chintala.
\newblock Pytorch: An imperative style, high-performance deep learning library,
  2019.
\newblock URL \url{https://arxiv.org/abs/1912.01703}.

\bibitem[Petroni et~al.(2019)Petroni, Rocktäschel, Lewis, Bakhtin, Wu, Miller,
  and Riedel]{petroni2019languagemodelsknowledgebases}
Fabio Petroni, Tim Rocktäschel, Patrick Lewis, Anton Bakhtin, Yuxiang Wu,
  Alexander~H. Miller, and Sebastian Riedel.
\newblock Language models as knowledge bases?, 2019.
\newblock URL \url{https://arxiv.org/abs/1909.01066}.

\bibitem[Radford et~al.(2021)Radford, Kim, Hallacy, Ramesh, Goh, Agarwal,
  Sastry, Askell, Mishkin, Clark, Krueger, and
  Sutskever]{radford2021learningtransferablevisualmodels}
Alec Radford, Jong~Wook Kim, Chris Hallacy, Aditya Ramesh, Gabriel Goh,
  Sandhini Agarwal, Girish Sastry, Amanda Askell, Pamela Mishkin, Jack Clark,
  Gretchen Krueger, and Ilya Sutskever.
\newblock Learning transferable visual models from natural language
  supervision, 2021.
\newblock URL \url{https://arxiv.org/abs/2103.00020}.

\bibitem[Rajaby~Faghihi et~al.(2021)Rajaby~Faghihi, Guo, Uszok, Nafar, and
  Kordjamshidi]{rajaby-faghihi-etal-2021-domiknows}
Hossein Rajaby~Faghihi, Quan Guo, Andrzej Uszok, Aliakbar Nafar, and Parisa
  Kordjamshidi.
\newblock {D}omi{K}now{S}: A library for integration of symbolic domain
  knowledge in deep learning.
\newblock In Heike Adel and Shuming Shi, editors, \emph{Proceedings of the 2021
  Conference on Empirical Methods in Natural Language Processing: System
  Demonstrations}, pages 231--241, Online and Punta Cana, Dominican Republic,
  November 2021. Association for Computational Linguistics.
\newblock \doi{10.18653/v1/2021.emnlp-demo.27}.
\newblock URL \url{https://aclanthology.org/2021.emnlp-demo.27/}.

\bibitem[Sathasivam(2011)]{sathasivam2011learning}
Saratha Sathasivam.
\newblock Learning rules comparison in neuro-symbolicintegration.
\newblock \emph{International Journal of Applied Physics and Mathematics},
  1\penalty0 (2):\penalty0 129, 2011.

\bibitem[Serafini and d’Avila Garcez(2016)]{serafini2016learning}
Luciano Serafini and Artur~S d’Avila Garcez.
\newblock Learning and reasoning with logic tensor networks.
\newblock In \emph{Conference of the italian association for artificial
  intelligence}, pages 334--348. Springer, 2016.

\bibitem[Shpilka et~al.(2010)Shpilka, Yehudayoff,
  et~al.]{shpilka2010arithmetic}
Amir Shpilka, Amir Yehudayoff, et~al.
\newblock Arithmetic circuits: A survey of recent results and open questions.
\newblock \emph{Foundations and Trends{\textregistered} in Theoretical Computer
  Science}, 5\penalty0 (3--4):\penalty0 207--388, 2010.

\bibitem[Silver et~al.(2016)Silver, Huang, Maddison, Guez, Sifre, van~den
  Driessche, Schrittwieser, Antonoglou, Panneershelvam, Lanctot, Dieleman,
  Grewe, Nham, Kalchbrenner, Sutskever, Lillicrap, Leach, Kavukcuoglu, Graepel,
  and Hassabis]{alphago}
David Silver, Aja Huang, Chris~J. Maddison, Arthur Guez, Laurent Sifre, George
  van~den Driessche, Julian Schrittwieser, Ioannis Antonoglou, Veda
  Panneershelvam, Marc Lanctot, Sander Dieleman, Dominik Grewe, John Nham, Nal
  Kalchbrenner, Ilya Sutskever, Timothy Lillicrap, Madeleine Leach, Koray
  Kavukcuoglu, Thore Graepel, and Demis Hassabis.
\newblock Mastering the game of go with deep neural networks and tree search.
\newblock \emph{Nature}, 529\penalty0 (7587):\penalty0 484--489, 2016.
\newblock \doi{10.1038/nature16961}.
\newblock URL \url{https://doi.org/10.1038/nature16961}.

\bibitem[Smolensky et~al.(2016)Smolensky, Lee, He, Yih, Gao, and Deng]{tensor}
Paul Smolensky, Moontae Lee, Xiaodong He, Wen-tau Yih, Jianfeng Gao, and
  li~Deng.
\newblock Basic reasoning with tensor product representations.
\newblock 01 2016.
\newblock \doi{10.48550/arXiv.1601.02745}.

\bibitem[Sun et~al.(2021)Sun, Sun, Han, and Zhou]{pmlr-v155-sun21a}
Jiankai Sun, Hao Sun, Tian Han, and Bolei Zhou.
\newblock Neuro-symbolic program search for autonomous driving decision module
  design.
\newblock In Jens Kober, Fabio Ramos, and Claire Tomlin, editors,
  \emph{Proceedings of the 2020 Conference on Robot Learning}, volume 155 of
  \emph{Proceedings of Machine Learning Research}, pages 21--30. PMLR, 16--18
  Nov 2021.
\newblock URL \url{https://proceedings.mlr.press/v155/sun21a.html}.

\bibitem[Tjong Kim~Sang and
  De~Meulder(2003)]{tjong-kim-sang-de-meulder-2003-introduction}
Erik~F. Tjong Kim~Sang and Fien De~Meulder.
\newblock Introduction to the {C}o{NLL}-2003 shared task: Language-independent
  named entity recognition.
\newblock In \emph{Proceedings of the Seventh Conference on Natural Language
  Learning at {HLT}-{NAACL} 2003}, pages 142--147, 2003.
\newblock URL \url{https://aclanthology.org/W03-0419/}.

\bibitem[Touvron et~al.(2023)Touvron, Martin, Stone, Albert, Almahairi, Babaei,
  Bashlykov, Batra, Bhargava, Bhosale, Bikel, Blecher, Ferrer, Chen, Cucurull,
  Esiobu, Fernandes, Fu, Fu, Fuller, Gao, Goswami, Goyal, Hartshorn, Hosseini,
  Hou, Inan, Kardas, Kerkez, Khabsa, Kloumann, Korenev, Koura, Lachaux, Lavril,
  Lee, Liskovich, Lu, Mao, Martinet, Mihaylov, Mishra, Molybog, Nie, Poulton,
  Reizenstein, Rungta, Saladi, Schelten, Silva, Smith, Subramanian, Tan, Tang,
  Taylor, Williams, Kuan, Xu, Yan, Zarov, Zhang, Fan, Kambadur, Narang,
  Rodriguez, Stojnic, Edunov, and Scialom]{touvron2023llama2openfoundation}
Hugo Touvron, Louis Martin, Kevin Stone, Peter Albert, Amjad Almahairi, Yasmine
  Babaei, Nikolay Bashlykov, Soumya Batra, Prajjwal Bhargava, Shruti Bhosale,
  Dan Bikel, Lukas Blecher, Cristian~Canton Ferrer, Moya Chen, Guillem
  Cucurull, David Esiobu, Jude Fernandes, Jeremy Fu, Wenyin Fu, Brian Fuller,
  Cynthia Gao, Vedanuj Goswami, Naman Goyal, Anthony Hartshorn, Saghar
  Hosseini, Rui Hou, Hakan Inan, Marcin Kardas, Viktor Kerkez, Madian Khabsa,
  Isabel Kloumann, Artem Korenev, Punit~Singh Koura, Marie-Anne Lachaux,
  Thibaut Lavril, Jenya Lee, Diana Liskovich, Yinghai Lu, Yuning Mao, Xavier
  Martinet, Todor Mihaylov, Pushkar Mishra, Igor Molybog, Yixin Nie, Andrew
  Poulton, Jeremy Reizenstein, Rashi Rungta, Kalyan Saladi, Alan Schelten, Ruan
  Silva, Eric~Michael Smith, Ranjan Subramanian, Xiaoqing~Ellen Tan, Binh Tang,
  Ross Taylor, Adina Williams, Jian~Xiang Kuan, Puxin Xu, Zheng Yan, Iliyan
  Zarov, Yuchen Zhang, Angela Fan, Melanie Kambadur, Sharan Narang, Aurelien
  Rodriguez, Robert Stojnic, Sergey Edunov, and Thomas Scialom.
\newblock Llama 2: Open foundation and fine-tuned chat models, 2023.
\newblock URL \url{https://arxiv.org/abs/2307.09288}.

\bibitem[{Van Hentenryck} et~al.(1992){Van Hentenryck}, Simonis, and
  Dincbas]{VANHENTENRYCK1992113}
Pascal {Van Hentenryck}, Helmut Simonis, and Mehmet Dincbas.
\newblock Constraint satisfaction using constraint logic programming.
\newblock \emph{Artificial Intelligence}, 58\penalty0 (1):\penalty0 113--159,
  1992.
\newblock ISSN 0004-3702.
\newblock \doi{https://doi.org/10.1016/0004-3702(92)90006-J}.
\newblock URL
  \url{https://www.sciencedirect.com/science/article/pii/000437029290006J}.

\bibitem[Weizenbaum(1966)]{10.1145/365153.365168}
Joseph Weizenbaum.
\newblock Eliza—a computer program for the study of natural language
  communication between man and machine.
\newblock \emph{Commun. ACM}, 9\penalty0 (1):\penalty0 36–45, January 1966.
\newblock ISSN 0001-0782.
\newblock \doi{10.1145/365153.365168}.
\newblock URL \url{https://doi.org/10.1145/365153.365168}.

\bibitem[Wong et~al.(2023)Wong, Grand, Lew, Goodman, Mansinghka, Andreas, and
  Tenenbaum]{wong2023wordmodelsworldmodels}
Lionel Wong, Gabriel Grand, Alexander~K. Lew, Noah~D. Goodman, Vikash~K.
  Mansinghka, Jacob Andreas, and Joshua~B. Tenenbaum.
\newblock From word models to world models: Translating from natural language
  to the probabilistic language of thought, 2023.
\newblock URL \url{https://arxiv.org/abs/2306.12672}.

\bibitem[Xu et~al.(2024{\natexlab{a}})Xu, Wu, Sun, Ren, Yuan, Yuan, Lin, Qiao,
  and Liu]{xu-etal-2024-symbol}
Fangzhi Xu, Zhiyong Wu, Qiushi Sun, Siyu Ren, Fei Yuan, Shuai Yuan, Qika Lin,
  Yu~Qiao, and Jun Liu.
\newblock Symbol-{LLM}: Towards foundational symbol-centric interface for large
  language models.
\newblock In Lun-Wei Ku, Andre Martins, and Vivek Srikumar, editors,
  \emph{Proceedings of the 62nd Annual Meeting of the Association for
  Computational Linguistics (Volume 1: Long Papers)}, pages 13091--13116,
  Bangkok, Thailand, August 2024{\natexlab{a}}. Association for Computational
  Linguistics.
\newblock \doi{10.18653/v1/2024.acl-long.707}.
\newblock URL \url{https://aclanthology.org/2024.acl-long.707/}.

\bibitem[Xu et~al.(2018)Xu, Zhang, Friedman, Liang, and Van~den
  Broeck]{pmlr-v80-xu18h}
Jingyi Xu, Zilu Zhang, Tal Friedman, Yitao Liang, and Guy Van~den Broeck.
\newblock A semantic loss function for deep learning with symbolic knowledge.
\newblock In Jennifer Dy and Andreas Krause, editors, \emph{Proceedings of the
  35th International Conference on Machine Learning}, volume~80 of
  \emph{Proceedings of Machine Learning Research}, pages 5502--5511. PMLR,
  10--15 Jul 2018.
\newblock URL \url{https://proceedings.mlr.press/v80/xu18h.html}.

\bibitem[Xu et~al.(2024{\natexlab{b}})Xu, Wang, Dong, Wu, Zhang, and
  Chang]{xu2024faithful}
Xinyu Xu, Pan Wang, Shusen Dong, Ningyu Wu, Yue Zhang, and Baobao Chang.
\newblock Faithful logical reasoning via symbolic chain-of-thought.
\newblock \emph{arXiv preprint arXiv:2405.18357}, 2024{\natexlab{b}}.
\newblock URL \url{https://arxiv.org/abs/2405.18357}.

\bibitem[Yang et~al.(2024)Yang, Chen, and Tam]{yang-etal-2024-arithmetic}
Xiaocheng Yang, Bingsen Chen, and Yik-Cheung Tam.
\newblock Arithmetic reasoning with {LLM}: {P}rolog generation {\&}
  permutation.
\newblock In Kevin Duh, Helena Gomez, and Steven Bethard, editors,
  \emph{Proceedings of the 2024 Conference of the North American Chapter of the
  Association for Computational Linguistics: Human Language Technologies
  (Volume 2: Short Papers)}, pages 699--710, Mexico City, Mexico, June 2024.
  Association for Computational Linguistics.
\newblock \doi{10.18653/v1/2024.naacl-short.61}.
\newblock URL \url{https://aclanthology.org/2024.naacl-short.61/}.

\bibitem[Yao et~al.(2025)Yao, Peng, Mao, and
  Luo]{yao2025exploringlargelanguagemodels}
Liang Yao, Jiazhen Peng, Chengsheng Mao, and Yuan Luo.
\newblock Exploring large language models for knowledge graph completion, 2025.
\newblock URL \url{https://arxiv.org/abs/2308.13916}.

\bibitem[Yi et~al.(2018)Yi, Wu, Gan, Torralba, Kohli, and
  Tenenbaum]{yi2018neural}
Kexin Yi, Jiajun Wu, Chuang Gan, Antonio Torralba, Pushmeet Kohli, and
  Joshua~B. Tenenbaum.
\newblock Neural-symbolic vqa: Disentangling reasoning from vision and language
  understanding.
\newblock In \emph{Advances in Neural Information Processing Systems}, pages
  1039--1050, 2018.

\bibitem[Zhang et~al.(2023)Zhang, Huang, Li, Naik, and
  Xing]{zhang2023improvedlogicalreasoninglanguage}
Hanlin Zhang, Jiani Huang, Ziyang Li, Mayur Naik, and Eric Xing.
\newblock Improved logical reasoning of language models via differentiable
  symbolic programming, 2023.
\newblock URL \url{https://arxiv.org/abs/2305.03742}.

\bibitem[Zheng et~al.(2024)Zheng, Lapata, and
  Pan]{zheng2024reliablellmsknowledgebases}
Danna Zheng, Mirella Lapata, and Jeff~Z. Pan.
\newblock How reliable are llms as knowledge bases? re-thinking facutality and
  consistency, 2024.
\newblock URL \url{https://arxiv.org/abs/2407.13578}.

\end{thebibliography}

\end{document}